\documentclass{article}

\usepackage{microtype}
\usepackage{graphicx}
\usepackage{booktabs} 
\usepackage{natbib}
\usepackage{amssymb}
\usepackage{amsmath}
\usepackage{subcaption}
\usepackage{enumitem}
\usepackage{mathtools}
\usepackage{multirow}
\usepackage{bbding}
\usepackage{wrapfig}
\usepackage[dvipsnames]{xcolor}
\usepackage{etoolbox}
\usepackage{tikz}

\usepackage{hyperref}

\newrobustcmd*{\mycircle}[1]{\tikz{\filldraw[draw=#1,fill=#1] (0,0) circle [radius=0.1cm];}}
\newrobustcmd*{\mytriangle}[1]{\tikz{\filldraw[draw=#1,fill=#1] (0,0) --
(0.2cm,0) -- (0.1cm,0.2cm);}}

\newcommand{\prelnppl}{24.5}
\newcommand{\postlnppl}{24.9}

\usepackage[accepted]{icml2021}

\icmltitlerunning{Stabilizing Equilibrium Models by Jacobian Regularization}

\begin{document}

\twocolumn[
\icmltitle{Stabilizing Equilibrium Models by Jacobian Regularization}



\icmlsetsymbol{equal}{*}

\begin{icmlauthorlist}
\icmlauthor{Shaojie Bai}{cmu}
\icmlauthor{Vladlen Koltun}{intel}
\icmlauthor{J. Zico Kolter}{cmu}
\end{icmlauthorlist}

\icmlaffiliation{cmu}{Carnegie Mellon University, Pittsburgh PA, USA}
\icmlaffiliation{intel}{Intel Labs, USA}

\icmlcorrespondingauthor{Shaojie Bai}{shaojieb@cs.cmu.edu}

\icmlkeywords{Machine Learning, ICML}

\vskip 0.3in
]



\printAffiliationsAndNotice{}  

\begin{abstract}
Deep equilibrium networks (DEQs) are a new class of models that eschews traditional depth in favor of finding the fixed point of a single non-linear layer. These models have been shown to achieve performance competitive with the state-of-the-art deep networks while using significantly less memory. Yet they are also slower, brittle to architectural choices, and introduce potential instability to the model. In this paper, we propose a regularization scheme for DEQ models that explicitly regularizes the Jacobian of the fixed-point update equations to stabilize the learning of equilibrium models. We show that this regularization adds only minimal computational cost, significantly stabilizes the fixed-point convergence in both forward and backward passes, and scales well to high-dimensional, realistic domains (e.g., WikiText-103 language modeling and ImageNet classification). Using this method, we demonstrate, for the first time, an implicit-depth model that runs with approximately the same speed and level of performance as popular conventional deep networks such as ResNet-101, while still maintaining the constant memory footprint and architectural simplicity of DEQs. Code is available \href{https://github.com/locuslab/deq}{here}.
\end{abstract}

\section{Introduction}
\label{sec:intro}

While conventional deep networks like ResNets~\citep{he2016deep} and Transformers~\citep{vaswani2017attention} rely on hierarchical layer stacking, the recently-proposed deep equilibrium networks (DEQs)~\citep{bai2019deep} directly model the ``infinite-depth'' representation of a single layer $f_\theta$ by solving for its fixed point (i.e., ``equilibrium'') $\mathbf{z}^\star$:
$$
\mathbf{z^\star} = f_\theta(\mathbf{z}^\star; \mathbf{x}),
$$
where $\mathbf{x}$ is the original input. Importantly, to train these models, one could directly differentiate through the final equilibrium $\mathbf{z}^\star$ by the implicit function theorem~\citep{krantz2012implicit}, irrespective of the method used to solve for this equilibrium in the forward pass. Therefore, like other implicit-depth architectures such as Neural ODEs~\citep{chen2018neural}, DEQs have the notable advantages that their forward passes can rely on any black-box root solvers (e.g., Newton, quasi-Newton, simplest forward iterations), and that their training only consumes $O(1)$ memory. With this formulation, prior works have managed to extend the DEQ framework for multiple large-scale applications, such as language modeling~\citep{bai2019deep} and large-scale image classification or segmentation~\citep{bai2020multiscale}.

However, these models suffer from a few issues. First, despite their memory efficiency, DEQs are also slower than conventional deep networks that achieve the same level of accuracy. Second, the number of iterations required to solve for the equilibrium quickly grows over the course of training, indicating a trend for approaching instability. Third, the DEQ model is sensitive to architectural choices, and sometimes even small modifications could break the model's stability of convergence. Some recent works have tackled this third issue by exploiting provably convergent layers via monotone operator splitting theories~\citep{winston2020monotone} and Lipschitz boundedness~\citep{revay2020lipschitz}. However, these structural solutions rely extensively on specific layer parameterizations, rendering DEQ models unscalable and even more inflexible.

In this paper, we first summarize and provide empirical evidence on all of these downsides of the equilibrium networks that have so far thwarted many from extending DEQs to both broader applications and more architectural variants. To address these issues, we further propose a regularization solution to improve on DEQ models' stability, efficiency and flexibility. Importantly, while prior DEQs adopted regularization methods direcly borrowed from explicit deep networks (e.g., recurrent dropout~\citep{gal2016dropout}), we introduce a simple and theoretically-motivated Jacobian regularization pursuant to DEQ models' implicitness. We will discuss in detail how this Jacobian regularization relates to the contractivity of DEQ's forward non-linear system and backward linear system, and is thus able to effectively stabilize not only forward but also backward dynamics of DEQ networks. There are two immediate benefits of the resulting stability in the dynamics. First, solving a DEQ requires far fewer iterations than before, which makes regularized DEQs significantly faster than their unregularized counterparts. Second, this model class becomes much less brittle to architectural variants that would otherwise break the DEQ.

We validate the proposed regularization by experiments on both toy-scale synthetic tasks and large-scale real datasets across domains: word-level language modeling on WikiText-103~\citep{merity2016pointer} and high-resolutional image classification on the full ImageNet dataset~\citep{deng2009imagenet}. Empirically, our regularized DEQs are generally 2x to 3x faster than prior DEQs, and can be accelerated to be as fast as explicit deep networks (e.g., ResNets and DenseNets). This is the first time that implicit models are accelerated to this level without sacrificing scalability, accuracy, or structural flexibility. With their $O(1)$ memory footprint, this further establishes implicit models as a strong competitor to explicit deep architectures.

\section{Background \& Related Work}

While explicit deep networks hierarchically stack layers to build a computation graph for their forward and backward propagations, implicit models~\citep{amos2017optnet,chen2018neural,elghaoui2019implicit,gould2019deep,bai2019deep} do not have a prescribed computation graph. Instead these models solve for a non-linear system. One example is the Neural ODE~\citep{chen2018neural}, which solves an initial-value ODE problem that involves a residual layer. Another example, which is the primary focus of our work, is the deep equilibrium network (DEQ)~\citep{bai2019deep}, which reduces the forward pass to a root-solving problem. In this section, we introduce the basics of DEQ models and the relevant threads of work, followed by a discussion of prior approaches to regularizing implicit models.

\subsection{Deep Equilibrium Models}
\label{subsec:implicit}

Given a layer/block $f_\theta$ (which may contain a few shallow sublayers) and an input $\mathbf{x}$, a deep equilibrium model aims to approximate an ``infinite-level'' layer stacking of the form $\mathbf{z}^{[i+1]} = f_\theta(\mathbf{z}^{[i]};\mathbf{x})$ (where $i=1,\dots,L$, with $L \rightarrow \infty$) by directly solving for its fixed-point representation:
$$
\mathbf{z}^\star = f_\theta(\mathbf{z}^\star; \mathbf{x}).
$$
One of the appealing properties of this fixed-point formulation is that one can implicitly differentiate through the equilibrium feature, without dependency on any intermediate activations in the forward pass. Formally, given a loss $\ell$, one can directly compute the gradient using the final output:
$$
\frac{\partial \ell}{\partial (\cdot)} = \frac{\partial \ell}{\partial \mathbf{z}^\star} \left(I - J_{f_\theta}(\mathbf{z}^\star) \right)^{-1} \frac{\partial f_\theta(\mathbf{z}^\star; \mathbf{x})}{\partial (\cdot)},
$$
where $J_{f_\theta}(\mathbf{z}^\star)$ is the Jacobian matrix at equilibrium $\mathbf{z}^\star$.

To solve for the equilibrium, \citet{bai2019deep} proposed to use Broyden's method~\citep{broyden1965class} to find the root of $f_\theta(\mathbf{z}^\star; \mathbf{x}) - \mathbf{z}^\star=0$; later works~\citep{winston2020monotone,revay2020lipschitz} and a recent tutorial~\citep{Duvenaud2020} have applied other algorithms, such as Peaceman-Rachford splitting~\citep{peaceman1955numerical} and Anderson acceleration~\citep{anderson1965iterative}.

Compared to Neural ODEs, deep equilibrium networks have been demonstrated to scale well to large and high-dimensional tasks, such as language modeling, ImageNet classification, and semantic segmentation~\citep{bai2019deep,bai2020multiscale}, and are thus more applicable to domains where deep learning has been traditionally successful. However, unlike ODE flows, DEQ networks do not have a unique trajectory, and are not guaranteed to converge. Thus recent works have also begun to examine the stability and other theoretical properties of DEQs. \citet{winston2020monotone} propose a monotone DEQ that has a unique fixed point. \citet{pabbaraju2021estimating,revay2020lipschitz} further study the Lipschitz boundedness of monotone DEQs. \citet{kawaguchi2021dynamics} analyze the gradient dynamics of a linearized version of DEQs. \citet{lu2021implicit} apply an invertible equilibrium model to generative modeling via normalizing flows.

\subsection{Regularizing Implicit Models}
\label{subsec:reg-implicit}

Just like explicit deep networks, implicit networks can overfit to the dataset; but additionally, they can also become unstable. For instance, Neural ODEs are essentially modeling infintesimal steps of a residual block~\citep{he2016deep,chang2017multi}, and~\citet{grathwohl2019ffjord} found that weight decay \& spectral normalization~\citep{miyato2018spectral} are useful (though expensive) in reducing the rapidly growing number of functional evaluations (NFEs) needed to solve for the ODE endpoint. On the other hand, large-scale DEQ networks~\citep{bai2019deep,bai2020multiscale} have adopted weight normalization~\citep{Salimans2016}, recurrent dropout~\citep{gal2016dropout}, and group normalization~\citep{wu2018group} for preventing overfitting and divergence. Nonetheless, all these methods are borrowed from explicit deep networks, where they have long been known to work well. They do not exploit the implicitness of implicit models.

\begin{figure*}[t]
\centering
\begin{subfigure}[b]{0.4\textwidth}
\includegraphics[width=\textwidth]{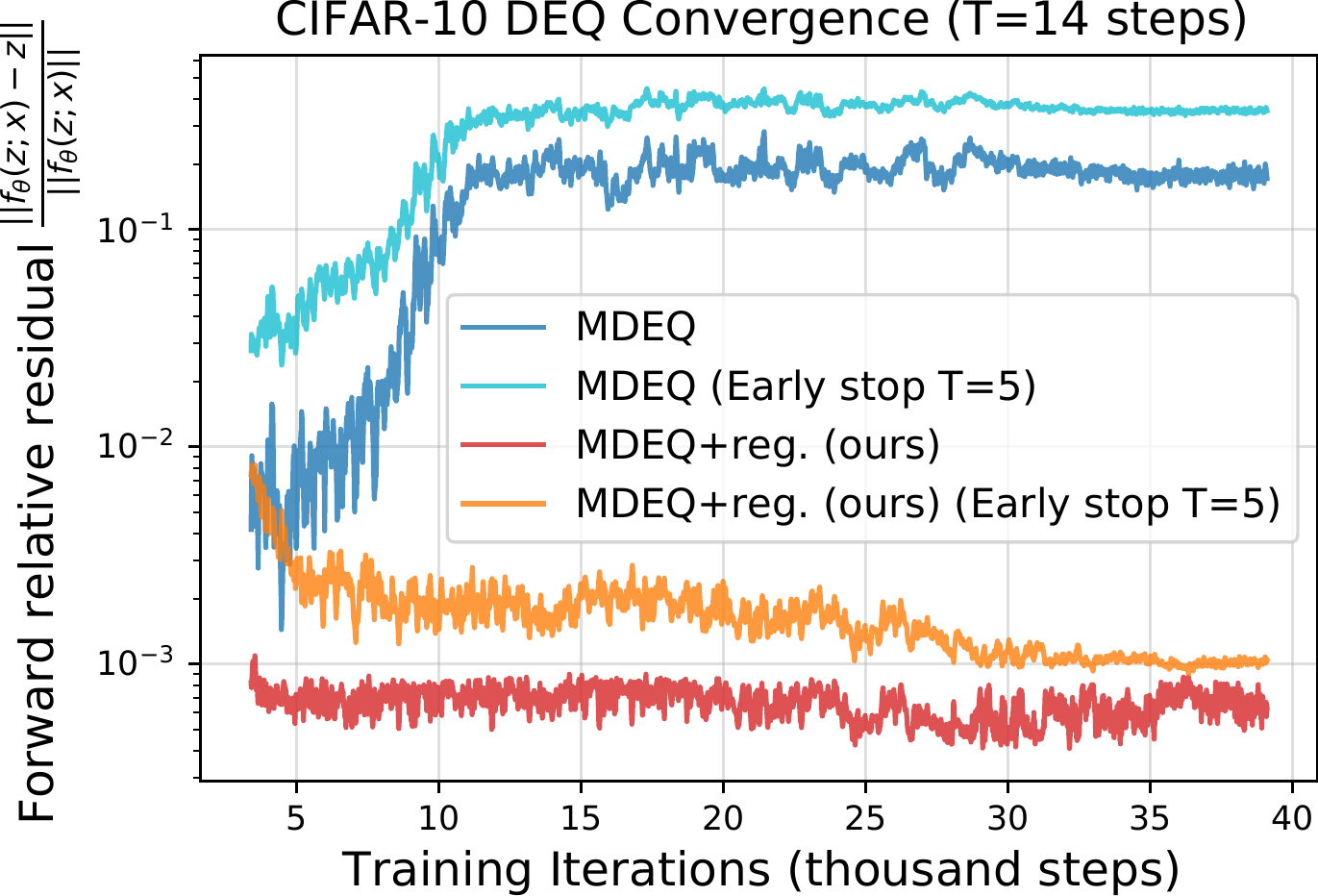}
\caption{Without regularizations, the relative residual of a DEQ's final output gets worse over training. Both models achieve roughly the same eventual level of accuracy.}
\label{subfig:cifar-convergence}
\end{subfigure}
~
\begin{subfigure}[b]{0.3\textwidth}
\includegraphics[width=\textwidth]{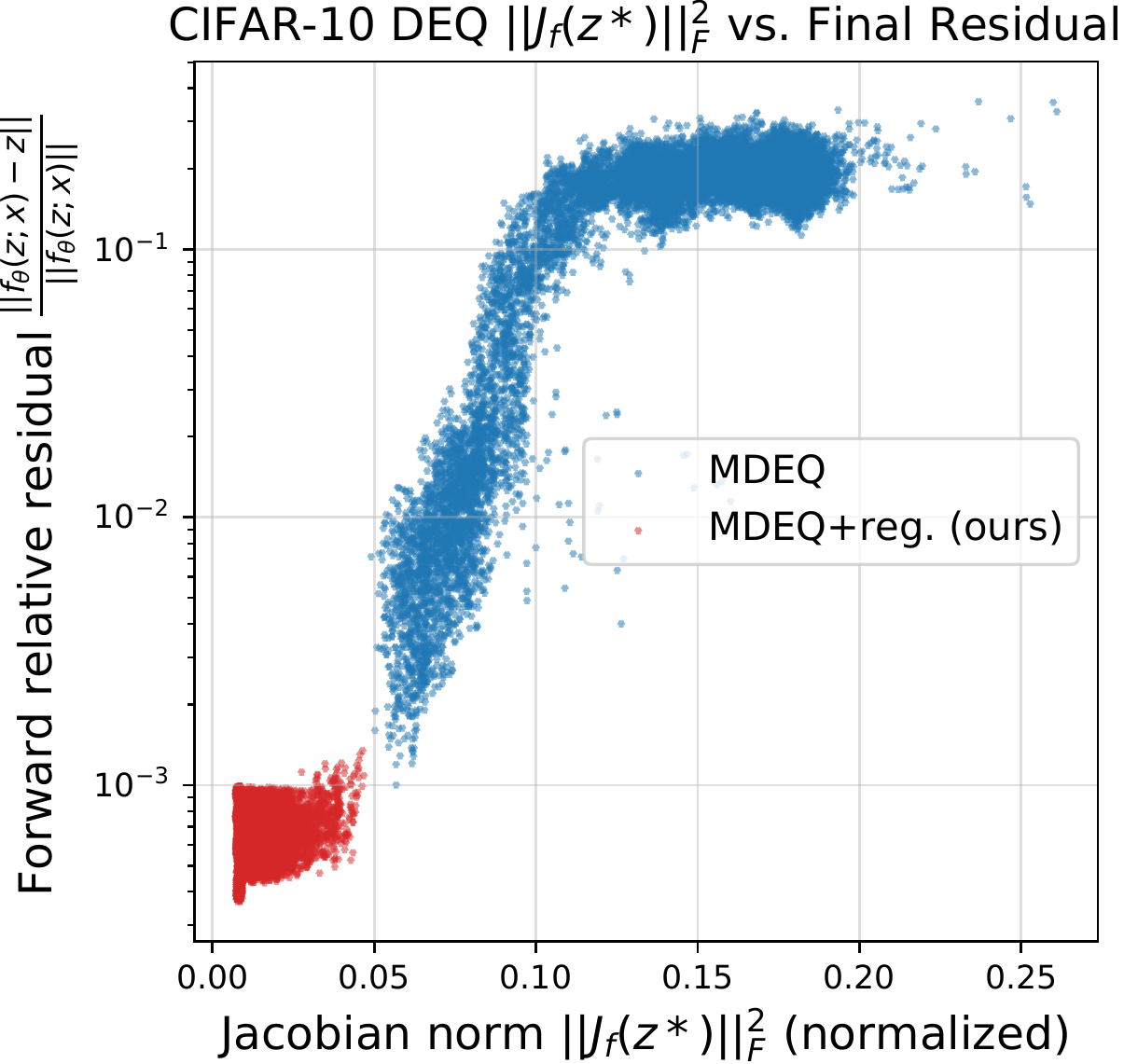}
\caption{In the same setting as Figure~\ref{subfig:cifar-convergence}, DEQ's convergence residual vs.\ Jacobian norm $\|J_f\|_F^2$.}
\label{subfig:jacobian-increase}
\end{subfigure}
~
\begin{subfigure}[b]{0.23\textwidth}
\includegraphics[width=\textwidth]{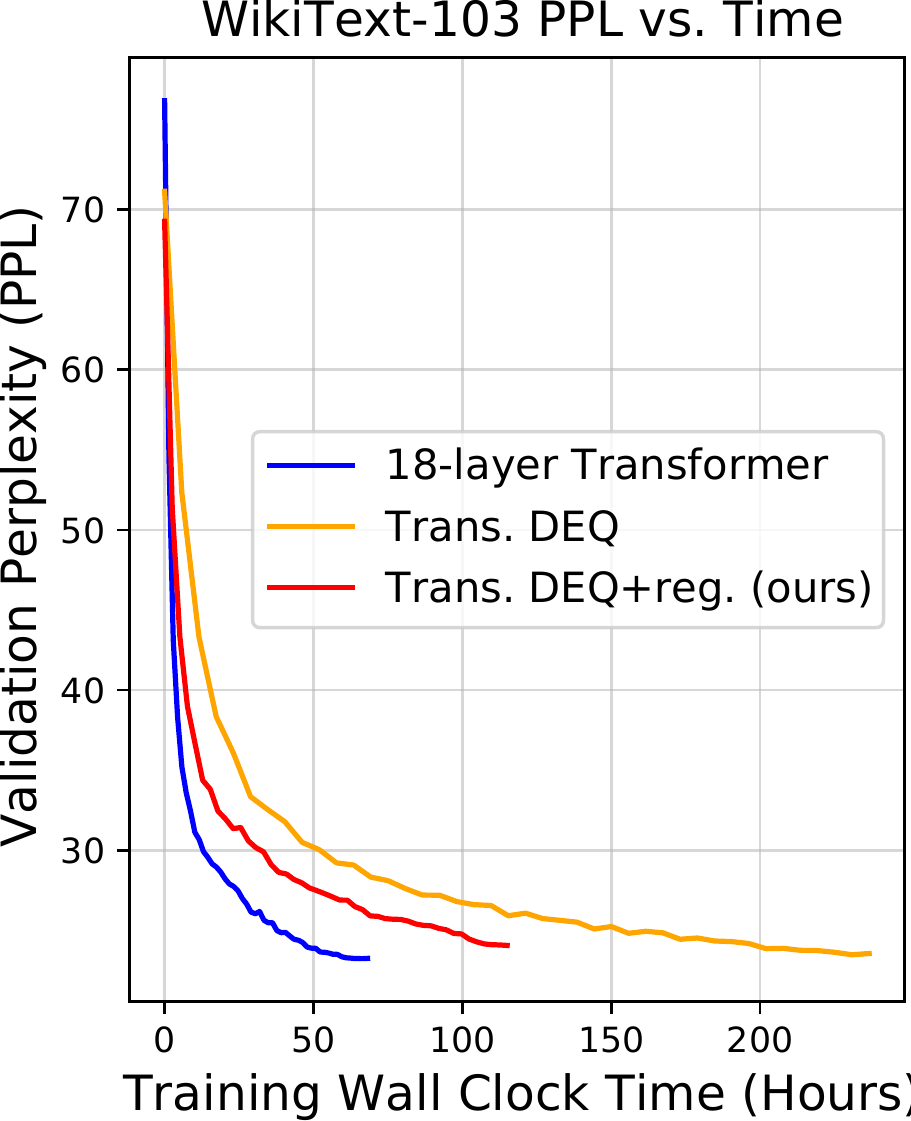}
\caption{Perplexity (ppl) of DEQs on WikiText-103 language modeling as a function of time.}
\label{subfig:wt103-convergence}
\end{subfigure}
\vspace{-.1in}
\caption{Visualizations of DEQs' instablity and inefficiency problems.}

\label{fig:problems1}
\vspace{-.12in}
\end{figure*}

More recently, a few different regularization methods have been introduced to specifically fix the numerous issues of the vanilla Neural ODE and continuous normalizing flow models, such as augmenting the hidden state~\citep{dupont2019augmented}, using hyper ODE solvers~\citep{poli2020hypersolvers}, and regularizing higher-order time derivatives~\citep{kelly2020learning}. These methods directly leverage the dynamical system view of Neural ODEs. However, due to the inherent challenge of solving high-dimensional ODEs, these accelerated Neural ODE models can still easily take $>100$ forward iterations even on MNIST classification~\citep{kelly2020learning}, and even more for their backward pass. In comparison, DEQs scale better to high-dimensional tasks (e.g., 25-30 iterations on ImageNet)~\citep{bai2020multiscale} and complex $f_\theta$ (e.g., a Transformer layer). But such extra complexities also make DEQ models harder to regularize; e.g., simply resorting to weight decay doesn't fix the instability of DEQs (see Section~\ref{subsec:ablative-limitation}). To the best of our knowledge, there has been almost no exploration of directly regularizing DEQ stability and convergence.

Our method is closely connected to the many prior works that study Jacobian/gradient regularization~\citep{drucker1992improving,novak2018sensitivity,hoffman2019robust,finlay2020train,linsley2020stable}, though these were also motivated differently. Specifically, \citet{sokolic2017robust,hoffman2019robust} regularized the input-output Jacobians of the entire (very deep) explicit classification networks to increase the prediction margin in a robust learning setting (and are thus expensive).~\citet{finlay2020train} was inspired by a kinetic energy view and possible overfitting of a training-time dynamical system. The method of~\citet{linsley2020stable} targeted (for a Jacobian $J$) a Lipschitzness level $\lambda$, used $\max_i (\mathbf{1}^\top J)_i$ to approximate the matrix 1-norm, and proposed loss $L=\|(\mathbf{1}^\top J - \lambda)^+\|_2$. Yet this approximation is in fact problematic, as it does not provably bound the spectral radius (i.e., stability) at all. For example, matrix
\[
J = \begin{bmatrix}2 & -2 \\ -2 & 2\end{bmatrix}
\]
has $L=0$ and yet an eigenvalue of 4 (we also empirically verify that this method does not help DEQ models, exactly due to this issue).

In contrast to these works, the key contributions of our paper are that (1) we provide a thorough discussion \& summary of various issues with DEQ models, and how ill-conditioned Jacobians are related to the forward/backward instabilities, via the new lens of fixed-point convergence; and (2) we demonstrate how regularizing the Jacobian of DEQs at the equilibrium point (i.e., the final output $\mathbf{z}^\star$) can provably bound the stability of the forward and backward convergences, thereby addressing these various problems. For example, our experiments show that we can significantly stabilize DEQs with new (and more unstable) architectural variants and accelerate DEQs to be nearly as fast as certain explicit architectures (e.g., we only need $\leq 6$ NFEs on CIFAR-10) on tasks across different scales and with comparable accuracy.

\section{DEQ Models and Their Discontents}
\label{sec:discontent}

Despite the DEQ models' success in some very challenging tasks, such as Cityscapes semantic segmentation~\citep{cordts2016cityscapes,bai2020multiscale}, these models suffer from multiple serious downsides. In this section, we provide a summary of some of these problems. While these issues directly lead to our subsequent discussion on the need for regularization (see Section~\ref{sec:regdeq}), we also believe such systematic discussion provides a useful overview for potential future research on further addressing these issues.

\subsection{Growing Instability during Training}
\label{subsec:grow-instability}

Although a DEQ network has no ``depth'', a relevant measure of computational efficiency is the number of function evaluations (NFEs) of the layer $f_\theta(\mathbf{z}; \mathbf{x})$ used by the iterative root solver (e.g., Broyden's method~\citep{broyden1965class}).

However, one common phenomenon to all prior works on DEQs is that the fixed points are increasingly harder to solve for over the course of model training. In other words, as a DEQ's performance gradually improves during training, the NFE required to converge to the same threshold $\varepsilon$ (e.g., $10^{-3}$) rapidly grows. This observation has been made on different instantiations of equilibrium networks, and regardless of whether the model is provably convergent or not (e.g.,~\citep{bai2019deep,winston2020monotone}, where a DEQ at the end of training can take $>3\times$ more iterations). Intuitively, such tendency to approach ``critical stability'' implicitly characterizes an inclination of the model to learn ``deeper'' networks; so it is unsurprising that unregularized training will keep driving it in this direction. But as a result, the dynamical system only becomes more and more brittle.  The existing way of ``addressing'' this is to circumvent it by setting a maximum NFE limit besides the $\varepsilon$-threshold; i.e., the solver stops either when 1) the residual is smaller than $\varepsilon$, or 2) it has run for a max number of steps $T$. This could be risky because as the convergence gets more unstable/critical, such a hard stop for the solver cannot guarantee that we are close enough to the fixed point. In the backward pass, for instance, we may consequently be training DEQs with very noisy gradients. A similar issue exists for Neural ODEs, though these cannot easily be hard-stopped like DEQs due to the need to accurately trace the flow to the endpoint.

\begin{figure}[t]
\centering
\begin{subfigure}[b]{0.225\textwidth}
\includegraphics[width=\textwidth]{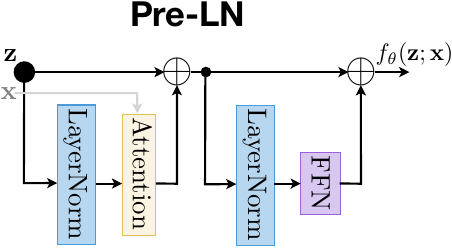}
\label{subfig:pre-ln}
\end{subfigure}
~
\begin{subfigure}[b]{0.235\textwidth}
\includegraphics[width=\textwidth]{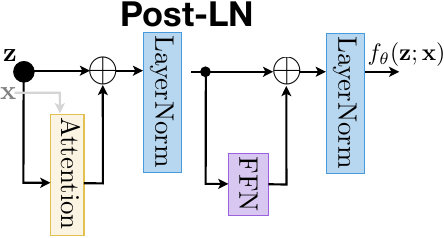}
\label{subfig:post-ln}
\end{subfigure}
\vspace{-.26in}
\caption{Pre- vs.\ post-LN DEQ-Transformer layer~\citep{xiong2020layer}. FFN is a 2-layer feed-forward block~\citep{vaswani2017attention}.}
\label{fig:pre-postln}
\vspace{-.18in}
\end{figure}

We illustrate this issue on CIFAR-10 classification in Fig.~\ref{subfig:cifar-convergence}. One can easily see that both forward and backward estimates of the fixed points gets increasingly worse with the training steps (and eventually plateaus in an unstable region where the model keeps yielding bad gradients). Such growing instability is also reflected empirically in the growth of Jacobian norm at equilibrium; i.e., \large$\left\|\frac{\partial f_\theta(\mathbf{z}^\star; \mathbf{x})}{\partial \mathbf{z}^\star} \right\|_F$\normalsize (see Figure~\ref{subfig:jacobian-increase}), which we discuss in Section~\ref{sec:regdeq}. Moreover, interestingly, while these plots might suggest simple regularizations like weight decay, we show later that weight decay often makes this stability issue worse for equilibrium networks, and even leads to divergence.

\subsection{Inefficiency Compared to Explicit Networks}

A direct ramification of the increase in iterations required (see Section~\ref{subsec:grow-instability}) is the significant increase in both training and inference time for DEQ models.


One advantage of DEQs noted by~\citet{bai2019deep} is that the forward trajectory need not strictly reach the equilibrium. Therefore in a certain sense, we could trade performance for efficiency by stopping at a ``good enough'' estimate of the equilibrium. However, due to the growing instability problem, this could still be increasingly costly. This causes the existing DEQs to be significantly slower than their explicit network counterparts of comparable size and performance. E.g., a DEQ-Transformer~\citep{bai2019deep} is about $3\times$ slower than a deep Transformer-XL~\citep{dai2018transformer}; a multiscale DEQ~\citep{bai2020multiscale} is over $4\times$ slower than ResNet-101 on ImageNet. Despite their memory efficiency, such slowdown is a roadblock to wider deployment of this class of models in practice. In Figure~\ref{subfig:wt103-convergence}, we visualize this slowdown on the validation set of WikiText-103 language modeling~\citep{merity2016pointer} (with comparable model sizes and number of training steps).

\begin{figure}[t]
\centering
\begin{subfigure}[b]{0.49\textwidth}
\includegraphics[width=\textwidth]{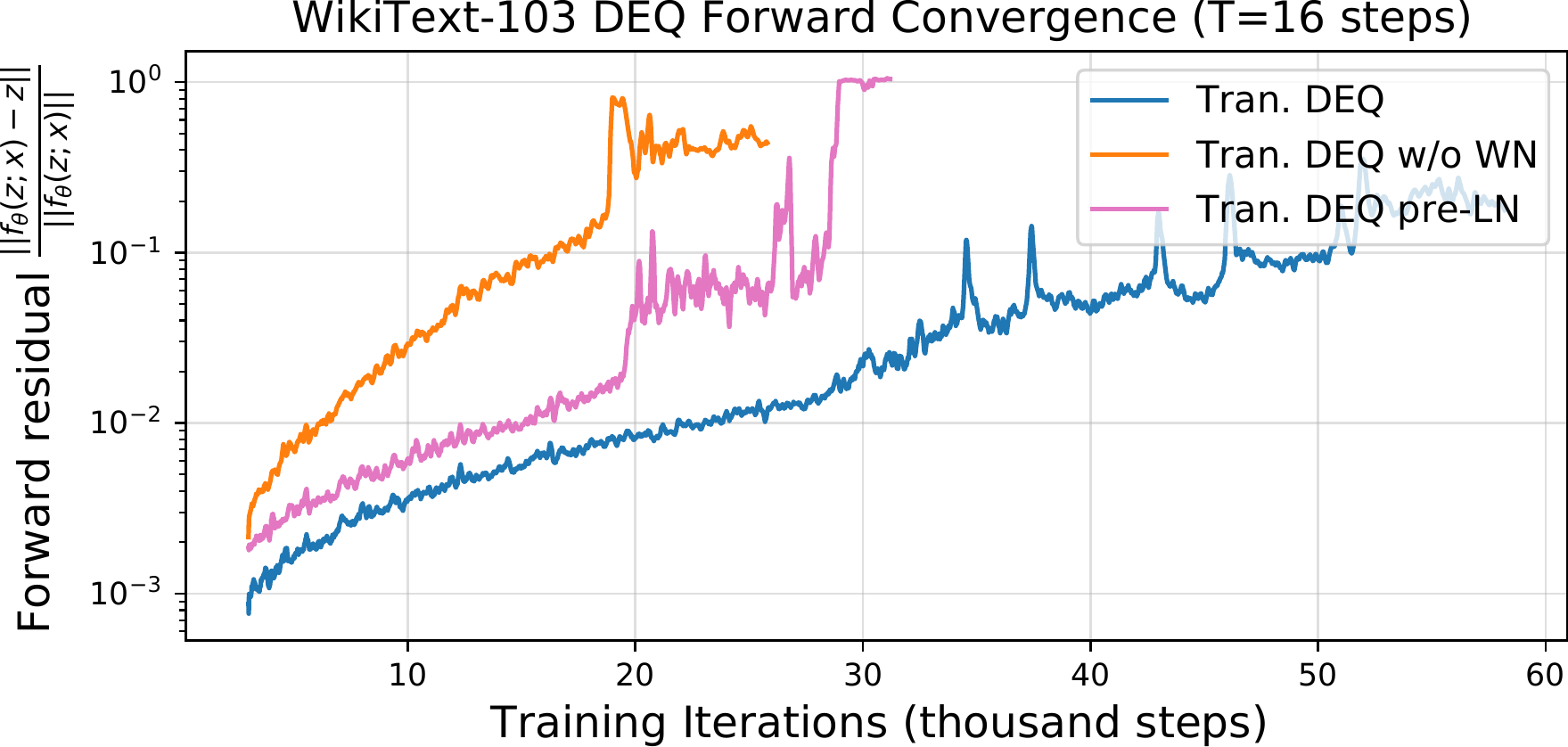}
\caption{Forward final objective of fixed-point convergence}
\label{subfig:architectural-choice-forward}
\vspace{.1in}
\end{subfigure}
~
\begin{subfigure}[b]{0.49\textwidth}
\includegraphics[width=\textwidth]{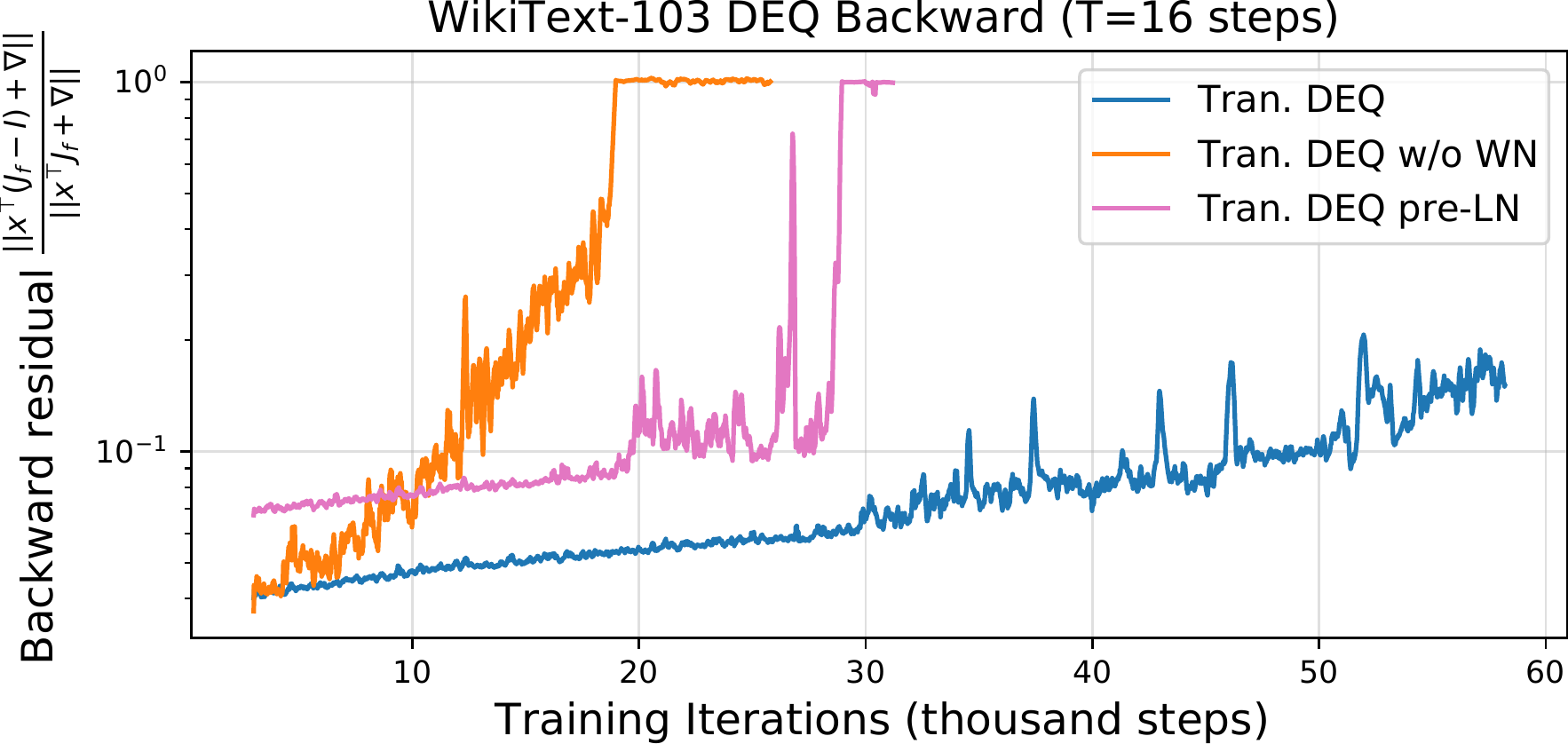}
\caption{Backward final objective of fixed-point convergence}
\label{subfig:architectural-choice-backward}
\end{subfigure}

\vspace{-.1in}
\caption{Comparing different architectural modifications of a DEQ-Transformer (first 60K steps). The DEQ networks are brittle: even slight modifications such as changing the whereabouts of LayerNorm (see Figure~\ref{fig:pre-postln}) or removing weight normalization can cause the model to quickly diverge during training.}
\label{fig:architectural-choice}
\vspace{-.2in}
\end{figure}

\subsection{Brittleness to Architectural Choices}
\label{subsec:architectural-choice}

The need to have a relatively stable DEQ in order to train it via the implicit function theorem also calls for more careful attention in designing the layer $f_\theta$. For example, the largest-scale DEQs~\citep{bai2019deep,bai2020multiscale} all had normalizations~\citep{Ba2016layer,wu2018group} at the end of the layer to constrain the output range. How important are these architectural choices? We demonstrate the brittleness of DEQs by ablative studies on the use of layer normalization (LN) or weight normalization (WN) in the DEQ-Transformer model on the large-scale WikiText-103 language modeling task.  Specifically, we compare the use of the two most popular Transformer layer designs in the DEQ framework: \emph{pre-LN} and \emph{post-LN}, which simply inserts the LN layers at different parts of the block (see Figure~\ref{fig:pre-postln}). These two settings have been extensively studied, used, and compared in the literature~\cite{liu2020understanding,xiong2020layer,vaswani2017attention,baevski2019adaptive,dosovitskiy2020image}.

The result is shown in Figure~\ref{fig:architectural-choice}. Without layer normalization at the end (\textcolor{Magenta}{magenta} line), the DEQ quickly diverges after 25K training iterations (reflected in both forward and backward divergences). Similarly, without weight normalization (\textcolor{orange}{orange} line), the model becomes unstable more quickly, with fixed-point solver collapse at around 18K iterations. The original DEQ-Transformer~\citep{bai2019deep} (\textcolor{blue}{blue} line in Figure~\ref{fig:architectural-choice}), although not diverged, still suffers from the same increased instability problem as described in Section~\ref{subsec:grow-instability}. These plots are strong indicators that while equilibrium networks work on large scales, they are also relatively inflexible, brittle, and reliant on meticulous architectural designs.

\subsection{The Hidden Cost of the Choice of Solver}
\label{subsec:solver-choice}

Although DEQ models enjoy constant memory consumption during training time and can use any black-box fixed point solvers in the forward and backward passes, a commonly neglected cost is that introduced by the choice of solver. For example, in Broyden's method~\citep{broyden1965class} which~\citet{bai2019deep,bai2020multiscale} used, the inverse Jacobian $J^{-1}$ is approximated by low-rank updates of the form ${J^{-1} \approx -I + \sum_{i=1}^n \mathbf{u}^{[n]}{\mathbf{v}^{[n]}}^\top = -I + UV^\top}$. As another example, Anderson mixing~\citep{anderson1965iterative} stores and uses the past $m$ iterations $(\mathbf{z}^{[n-1]}, \dots, \mathbf{z}^{[n-m]})$. In most such cases, even storing these updates or past steps can be expensive. Moreover, since we depend on the same DEQ solvers also at inference time, we need to spend this same memory cost even when when the trained model is served~-- which conventional deep networks can avoid. We note that this cost depends strongly on the solver; for example, the simplest iterative ``solver'' $\mathbf{z}^{[i+1]} = f_\theta(\mathbf{z}^{[i]}; \mathbf{x})$ wouldn't have any memory cost, but suffers from bad convergence. This issue also highlights the value of faster and stabler convergence, which entails less memory storage overall (e.g., fewer Broyden steps).

\section{Regularizing the Jacobian of DEQs}
\label{sec:regdeq}

We hypothesize that one of the fundamental factors contributing to some of the problems discussed in Section~\ref{sec:discontent} is that DEQ models' conditioning is not properly regularized during training. Such trend for DEQ models to go unstable is reflected in Figures~\ref{subfig:cifar-convergence} and~\ref{subfig:jacobian-increase}, where increasing training steps leads to monotonically growing residual difference and the Jacobian norm at the equilibrium. We now describe how the Jacobian is related to the stability of equilibrium networks' forward and backward passes, and then harness this relationship to stabilize and accelerate DEQs.

\subsection{The DEQ Jacobian}
\label{subsec:jacobian-and-deq}

\begin{figure}[t]
\centering
\includegraphics[width=0.49\textwidth]{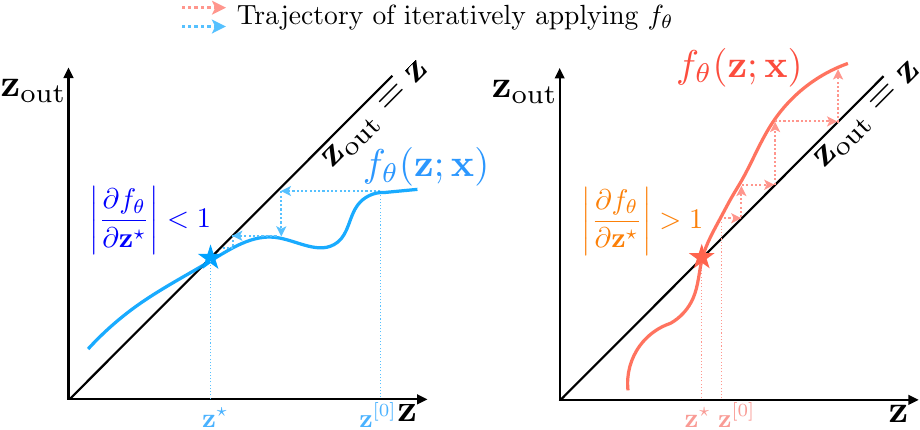}
\caption{Left: when the slope is less than 1, even the simplest iterative application of $f_\theta$ converges. Right: when slope $>1$, the iterative approach may diverge or oscillate, but the fixed point still exists and can be solved for.}
\vspace{-.1in}
\label{fig:op-norm}
\end{figure}

We first recall that the forward pass of a DEQ network aims to solve for the fixed-point representation $\mathbf{z}^\star$ of a layer $f_\theta(\cdot; \mathbf{x})$; i.e., $\mathbf{z}^\star = f_\theta(\mathbf{z}^\star)$. Then in the backward pass, one can differentiate directly through the equilibrium $\mathbf{z}^\star$ by
\begin{equation}
\label{eq:deq-backward}
\frac{\partial \ell}{\partial (\cdot)} = \underbrace{\frac{\partial \ell}{\partial \mathbf{z}^\star} \left(I - J_{f_\theta}(\mathbf{z}^\star) \right)^{-1}}_{\mathbf{u}^\top} \frac{\partial f_\theta(\mathbf{z}^\star; \mathbf{x})}{\partial (\cdot)}.
\end{equation}
However, because the scale of $J_{f_\theta}$ can be prohibitively large and the inverse is costly to compute, we usually compute the $\mathbf{u}^\top$ term in Eq.~\ref{eq:deq-backward} by solving the following linear fixed-point system that depends on the final Jacobian:
\begin{equation}
\label{eq:deq-backward-linear}
\mathbf{u}^\top = \mathbf{u}^\top J_{f_\theta}(\mathbf{z}^\star) + \frac{\partial \ell}{\partial \mathbf{z}^\star}.
\end{equation}

Consider the spectral radius of the Jacobian $J_{f_\theta} \in \mathbb{R}^{d \times d}$ at the equilibrium:
\begin{align*}
\rho(J_{f_\theta}(\mathbf{z}^\star)) = \rho(J_{f_\theta}(\mathbf{z}^\star)^\top) = \max(|\lambda_1|, \dots, |\lambda_d|),
\end{align*}
where $\lambda_i$s are eigenvalues. In both the forward and backward passes, this spectral radius directly affects how stable the convergence to the fixed point $\mathbf{z}^\star$ could be in its neighborhood. For instance, in the extreme case where we have a contractive $\rho(J_{f_\theta}) < 1$, by Lyapunov linearization theorem even the simplest iterative calls to $f_\theta(\mathbf{z})$ (in forward, assuming good initial estimate) or $g(\mathbf{u}) = \mathbf{u}^\top J_{f_\theta}(\mathbf{z}^\star) + \frac{\partial \ell}{\partial \mathbf{z}^\star}$ (in backward) could converge uniquely, even without advanced solvers. The linear system~\eqref{eq:deq-backward-linear}, in particular, would enjoy global asymptotic stability. However in practice, we don't always, and probably shouldn't, require such a strong contractivity on the dynamical system, which might significantly limit the representational capacity of the model. For example, as shown in Figure~\ref{fig:op-norm}, a fixed point can exist even if $\rho(J_{f_\theta}) > 1$, (the curve slope in 2D); and we are still able to solve for them using the much stronger root solvers (e.g., Newton or quasi-Newton) than these simplest iterative stackings, which could oscillate or diverge.

\subsection{Jacobian Regularization}
\label{subsec:jacobian-regularization}

These connections between $J_{f_\theta}(\mathbf{z}^\star)$ (which characterizes the shape of the transformation $f_\theta$ around $\mathbf{z}^\star$) and the forward/backward pass dynamics of DEQs motivate us to append a soft and auxiliary Jacobian term $\rho(J_{f_\theta}(\mathbf{z}^\star))$ to the training objective in order to regularize the model's conditioning. One way of doing this is by spectral normalization, essentially constraining $\sigma(J_{f_\theta}) = \max_{\|\mathbf{v}\|\leq 1} \|J_{f_\theta} \mathbf{v}\|_2$. However, explicitly writing out the huge Jacobian and then decomposing it (e.g., by SVD) can be computationally prohibitive, and~\citet{miyato2018spectral} proposes to use the power method~\citep{mises1929praktische} to speed up this estimation on GANs. But in the context of DEQs, even power iterations are too expensive due to the successive vector-Jacobian product computations needed. Instead, we propose to regularize the Jacobian through its Frobenius norm since
$$
\rho(J_{f_\theta}) \leq \sigma(J_{f_\theta}) \leq \sqrt{\text{tr}(J_{f_\theta} J_{f_\theta}^\top)} = \|J_{f_\theta}\|_F.
$$
Importantly, $\|J_{f_\theta}\|_F$ can be approximated via various unbiased estimators~\citep{hutchinson1989stochastic,ubaru2017fast,meyer2021hutch}. We adopt the classical Hutchinson estimator~\citep{hutchinson1989stochastic}; formally, for $J_{f_\theta} \in \mathbb{R}^{d \times d}$,
\begin{equation}
\vspace{-.02in}
\label{eq:hutchinson}
\text{tr}(J_{f_\theta} J_{f_\theta}^\top) = \mathbb{E}_{\epsilon \in \mathcal{N}(0,I_d)} [\|\epsilon^\top J_{f_\theta}\|_2^2],
\vspace{-.02in}
\end{equation}
which we can approximate by Monte-Carlo estimation (i.e., sampling $M$ i.i.d. $\epsilon_i \in \mathcal{N}(0, I_d)$). Specifically, prior works~\citep{avron2011randomized,roosta2015improved} have established that the relative error of this estimation diminishes with $M^{-\frac{1}{2}}$; and if we compute the mean estimation over a mini-batch size $B$, the overall relative error with respect to $\mathbb{E}_{\mathbf{x} \sim p(\mathbf{x}), \epsilon \in \mathcal{N}(0,I_d)}[\|\epsilon^\top J_{f_\theta}\|_2^2]$ is expected to further diminished by a factor of $B^{-\frac{1}{2}}$~\citep{hoffman2019robust}.

Indeed, empirically, we find that $M = 1$ already works well since we use relatively large batch sizes. Since our backward iterations already involved computing multiple vector-Jacobian products $\mathbf{u}^\top J_{f_\theta}$ (see Eq.~\eqref{eq:deq-backward-linear}), computing Eq.~\eqref{eq:hutchinson} only adds a cost equivalent to that of $M = 1$ backward steps. The eventual training objective is thus
\begin{align}
\label{eq:objective}
\resizebox{.435\textwidth}{!}{$
\mathcal{L}_\text{total}(\mathbf{z}^\star) = \mathcal{L}_\text{orig}(\mathbf{z}^\star) + \gamma \frac{\|\epsilon^\top J_{f_\theta}(\mathbf{z}^\star)\|_2^2}{d}, \ \ \epsilon \in \mathcal{N}(0,I_d)
$}
\end{align}
As we observed in Figure~\ref{subfig:cifar-convergence}, without regularization, a DEQ model that stops after a fixed number $T$ of solver iterations exhibits increasingly poor convergence, accompanied by a growing $\|J_{f_\theta}\|_F$ at these fixed points that empirically signals the growing instability. Therefore, by constraining the Jacobian's Frobenius norm, we encourage DEQs to optimize for stabler and simpler dynamics whose fixed points are easier to solve for.

\subsection{Memory Considerations}
\label{subsec:discussion}

Although the loss objective~\eqref{eq:objective} only adds minimal computation cost, the need to backpropate through $\|\epsilon^\top J_{f_\theta}\|_2^2$ means we also spend more memory during training to store the computation graph of this vector-Jacobian product. But at the same time, our hidden memory cost due to the solver choice is smaller (e.g., Broyden's method; see Section~\ref{subsec:solver-choice}) as we can lower the number of iterations. As a result, empirically we notice a roughly 30\% net growth in memory consumption compared to the unregularized DEQs at training (and thus saving about 50\% memory compared to explicit deep networks). The regularized DEQ still consumes $O(1)$ memory relative to the ``depth'' of the model, as the backpropagation depends only on $\mathbf{z}^\star$.

\section{Experiments}
\label{sec:experiments}

\begin{figure*}[t]
\centering
\begin{subfigure}[c]{0.335\textwidth}
\includegraphics[width=\textwidth]{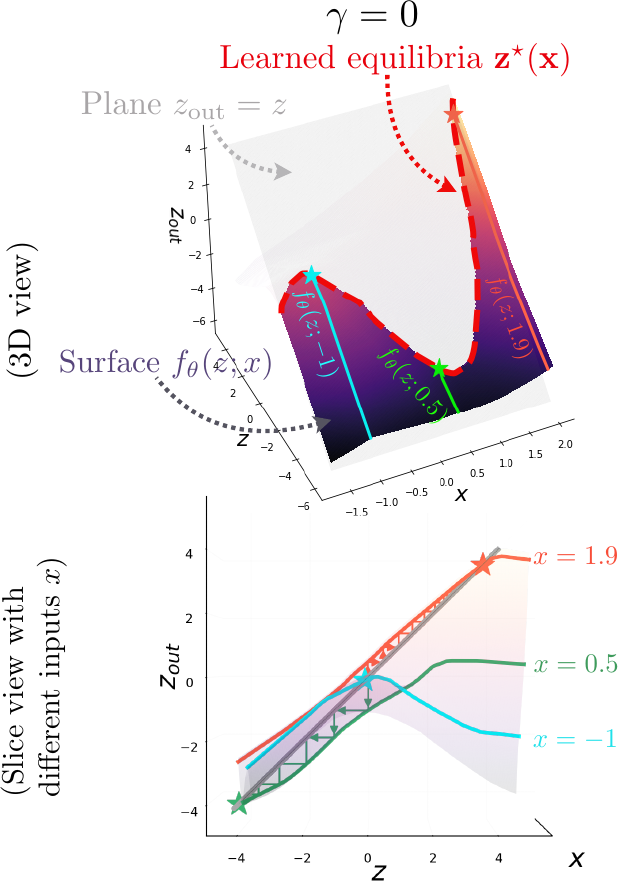}
\caption{}
\label{subfig:unreg_both}
\end{subfigure}
~
~
\begin{subfigure}[c]{0.225\textwidth}
\includegraphics[width=\textwidth]{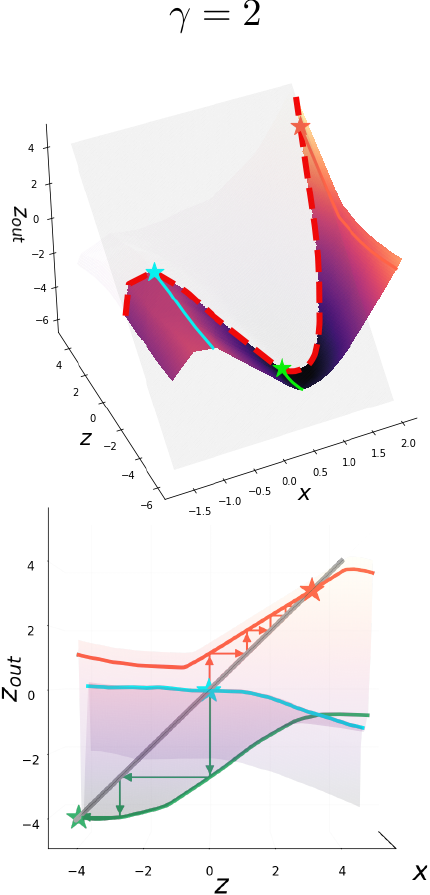}
\caption{}
\label{subfig:reg2_both}
\end{subfigure}
~
~
\begin{subfigure}[c]{0.225\textwidth}
\includegraphics[width=\textwidth]{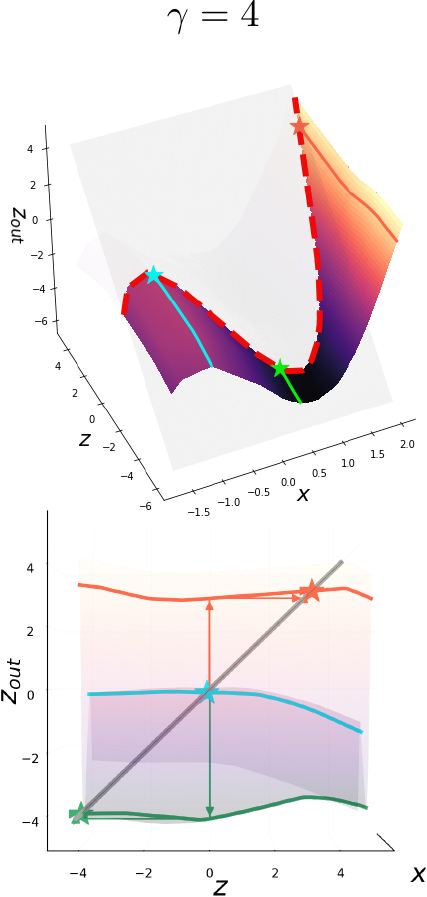}
\caption{}
\label{subfig:reg4_both}
\end{subfigure}
~
\begin{subfigure}[c]{0.15\textwidth}
\includegraphics[width=\textwidth]{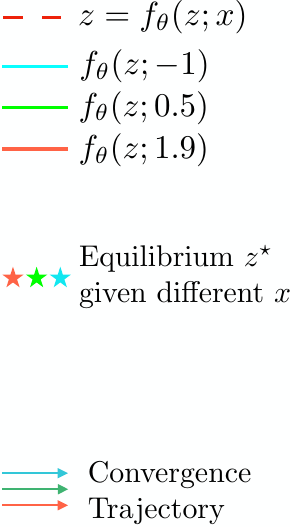}
\end{subfigure}

\vspace{-.12in}
\caption{\textbf{Top:} the surface of the $f_\theta(\mathbf{z};\mathbf{x})$ layer, and the eventual learned equilibria $z^\star(x)$ as a function of $x$. As $\gamma$ grows, the surface is ``lifted up'' and becomes flat in the $z$-direction. \textbf{Bottom:} each unique input $x$ defines a slice of the surface, and we perform fixed-point solving on this slice; larger $\gamma$ values flatten the curve and significantly accelerate the convergence to equilibrium.}
\label{fig:synthetic}
\vspace{-.2in}
\end{figure*}

We validate the proposed regularization of DEQ models on multiple fronts. First, we visualize the effect of the proposed Jacobian regularization on a tiny DEQ trained on a synthetic 1D dataset. Second, importantly, we focus on how our method alleviates some of the core problems with DEQs outlined in Section~\ref{sec:discontent}. Then we show that our method scales to challenging high-dimensional tasks: word-level language modeling with the WikiText-103 dataset~\citep{merity2016pointer} and image classification with CIFAR-10 and ImageNet~\citep{deng2009imagenet}. We specifically compare our model with both prior DEQ networks and competitive explicit models (e.g., ResNet-101, Transformers), in terms of both efficiency (in space and time) and performance. We also explore how Jacobian regularization helps stabilize DEQs over a wider range of architectural choices. Lastly, we perform some ablative studies.

The set of tasks used in our experiment is built directly on top of~\citet{bai2019deep,bai2020multiscale}. As we found the Jacoabian regularization could sometimes hurt performance (see Sec.~\ref{subsec:vision-tasks}), we only apply the proposed loss stochastically with a probability $p$, and gradually increase this $p$ or the regularization strength $\gamma$ (see Eq.~\eqref{eq:objective}) over training steps. We also use cosine learning rate schedule~\citep{loshchilov2017sgdr} for all tasks, including the synthetic one. The memory and speeds reported are benchmarked across different models on the same setting (e.g., same batch size, sequence length, number of steps, etc.) with the same GPU. We provide more details regarding the tasks, hyperparameters, datasets, and hardware in Appendix~\ref{app:experiment}, and extra experimental results in Appendix~\ref{app:extra-results}. Our code and pretrained models are provided \href{https://github.com/locuslab/deq}{here}.

\subsection{Visualization with Synthetic Data}

We start by empirically verifying the validity of the approach and visualizing its effect on a synthetic dataset. We generated 5096 scalar data pairs $(x,y)$ using function $y = h(x) = \frac{3}{2}x^3+x^2-5x+2\sin(x)-3 + \delta$ (where $\delta \in \mathcal{N}(0,0.05)$), and split them into 4096 and 1000 training and validation samples, respectively. We then train a tiny DEQ with 200 parameters with the following structure:
$$
f_\theta(\mathbf{z}; \mathbf{x}) = W_2^\top \ \text{ReLU}(W_1\mathbf{z} + U\mathbf{x} + b), \qquad \hat{y}=\mathbf{z}^\star
$$
where we used $\mathbf{z}, \mathbf{x} \in \mathbb{R}$ and $W_1, W_2, U \in \mathbb{R}^{50 \times 1}$. The visualizations of the effect of the Jacobian regularization, with different weights $\gamma$, are shown in Figure~\ref{fig:synthetic}.  In particular, each input $x$ defines a slice (i.e., cross-section) of the 3D surface $z_\text{out}=f_\theta(z; x)$; for example, layer $f_\theta(z; x)$ when input $x=-1$ is highlighted in \textcolor{ProcessBlue}{blue}. After training, all three settings succesfully learned the (almost) identical equilibrium function $z^\star(x)$ (highlighted by the \textcolor{red}{red} dashed line) that perfectly fits the target function $h(x)$; but note that surfaces of $f_\theta$ with $\gamma =2,4$ are ``lifted up'' significantly compared to the unregularized ($\gamma=0$) DEQ, which has a steep slope (i.e., large spectral radius in 2D). This slope slows down the fixed-point convergence, as reflected by the zigzag patterns in lower Figure~\ref{subfig:unreg_both}. In contrast, the convergences for the $\gamma>0$ cases are much faster, and larger $\gamma$ typically yields flatter surfaces around the equilibrium point.

\subsection{WikiText-103 Language Modeling}

\begin{table}[tb]
\caption{Evaluation on WikiText-103. PPL stands for Perplexity. All Transformer models are trained for 250K steps. $t_\text{train}$ stands for relative training time. \textbf{JR} stands for Jacobian regularization. NFEs are measured at inference time. $^\dagger$ indicates unregularized model hard-stopped at inference time.}
\label{table:wt103}
\centering
\def\arraystretch{1.04}
\vspace{2mm}
\resizebox{.49\textwidth}{!}{
\begin{tabular}{c|cccc}
\toprule
 & Size & PPL & NFEs & $t_\text{train}$ \\
 \midrule
 AWD-QRNN~\citep{bradbury2016quasi} & 159M & 33.0 & - & -\\
 Rel. Memory Core~\citep{santoro2018relational} & 195M & 31.6 & - & - \\
 18L-Transformer-XL~\citep{dai2018transformer} & 110M & 24.1 & - & $1 \times$  \\
 \midrule
 DEQ-Trans. (Pre-LN)~\citep{bai2019deep} & 98M & \textcolor{purple}{[div.]} & 30 & $3.1\times$ \\
 DEQ-Trans. (Post-LN)~\citep{bai2019deep} & 98M & 24.0 & 30 & $3.1\times$ \\
 DEQ-Trans. (Post-LN) \emph{early stopped}$^\dagger$ & 98M & 29.2 & 12$^\dagger$ & $3.1\times$ \\
 \textbf{DEQ-Trans. (Pre-LN) + JR (ours)} & 98M & \prelnppl & \textbf{14} & $1.6\times$ \\
 \textbf{DEQ-Trans. (Post-LN) + JR (ours)} & 98M & \postlnppl & \textbf{12} & $1.5\times$ \\
\bottomrule
\end{tabular}}
\vspace{-.15in}
\end{table}

One of the very first successes of large-scale DEQs was its Transformer instantiation~\citep{bai2019deep}, which uses a multi-head self-attention~\citep{vaswani2017attention} layer as the underlying $f_\theta(\mathbf{z};\mathbf{x})$ function. Although a DEQ-Transformer is able to perform competitively with a deep Transformer-XL~\citep{dai2018transformer} in terms of test perplexity, and consumes 60-70\% less memory, it is also much slower (about $3\times$; see Figure~\ref{subfig:wt103-convergence}) and borders on instability. In Table~\ref{table:wt103}, we demonstrate how the Jacobian regularization alleviates this. Compared to the original DEQ models, there are two major improvements. First, we significantly reduce the NFEs required for DEQ-Transformer models while maintaining competitive accuracy. Using the Transformer-XL as a time benchmark ($1\times$), the speed of a DEQ-Transformer is significantly accelerated: training time goes from $3.1\times$ to $1.5\times$. Second, the regularized DEQ model is more flexible with architectural choices. Whereas a Pre-LN DEQ-Transformer (see Figure~\ref{fig:pre-postln}) quickly diverges in training even in the presence of a large NFE threshold, the Jacobian regularization resolves this issue and stabilizes the forward/backward convergences consistently (see Figure~\ref{fig:architectural-choice} and Table~\ref{table:wt103}), eventually reaching \prelnppl~perplexity. Moreover, while we can early-stop a well-trained unregularized DEQ model at inference time, it hurts generalization performance significantly (e.g., 29.2 ppl with 12 NFEs). Similarly, we find training with NFEs $<30$ leads to increasingly bad generalization performance, and when NFEs drops below $20$, model training frequently diverge as a result of extremely noisy gradients. We provide more comprehensive results in Table~\ref{app-table:wt103} in the Appendix.

Like DEQs, the regularized DEQs are memory efficient, consuming about 45\% less training memory than Transformer-XL. Moreover, we find the Jacobian-regularized DEQs reduce over 50\% memory consumption of the original DEQs at inference time (when both using Broyden's method) due to faster/stabler convergence, suggesting its effectiveness in addressing the hidden solver cost issue discussed in Sec.~\ref{subsec:solver-choice}.

\subsection{CIFAR-10 and ImageNet Classification}
\label{subsec:vision-tasks}

\begin{table}[tb]
\caption{Results on CIFAR-10 and ImageNet classfication. The CIFAR-10 accuracy standard deviation is calculated with 5 runs. \textbf{JR} stands for Jacobian regularization. $^\dagger$ indicates unregularized model hard-stopped at inference time.}
\label{table:imagenet}
\centering
\def\arraystretch{1.04}
\vspace{2mm}
\resizebox{.49\textwidth}{!}{
\begin{tabular}{c|ccc}
\toprule
\multicolumn{4}{c}{CIFAR-10 classification} \\
 & Size & Accuracy & NFEs \\
 \midrule
 ResNet-18~\citep{he2016deep} & 10M & 93.0 ($\pm$ 0.1)\% & -\\
 ResNet-101~\citep{he2016deep} & 40M & 93.8 ($\pm$ 0.3)\% & -\\
 DenseNet-121~\citep{huang2017densely} & 8M & 95.0 ($\pm 0.1$)\% & - \\
 monotone DEQ~\citep{winston2020monotone} & 1M & 89.4 ($\pm$ 0.2)\% & 24\\
 MDEQ~\citep{bai2020multiscale} & 10M & 93.6 ($\pm$ 0.2)\% & 17 \\
 MDEQ \emph{early stopped}$^\dagger$ & 10M & 89.1\% & 6$^\dagger$ \\
 \textbf{MDEQ + JR (ours)}~\citep{bai2020multiscale} & 10M & 93.1 ($\pm$ 0.3)\% & \textbf{6} \\
 \midrule
 \multicolumn{4}{c}{(Full) ImageNet classification} \\
  & Size & Top-1 Acc. & NFEs \\
 \midrule
 ResNet-18~\citep{he2016deep} & 13M & 70.2\% & - \\
 Inception-V2~\citep{ioffe2015batch} & 12M & 74.8\% & - \\
 ResNet-50~\citep{he2016deep} & 26M & 75.1\% & - \\
 ResNet-101~\citep{he2016deep} & 52M & 77.1\% & - \\
 DenseNet-264~\citep{huang2017densely} & 74M & 79.7\% & - \\
 MDEQ-small~\citep{bai2020multiscale} & 18M & 75.4\% & 27 \\
 MDEQ-large~\citep{bai2020multiscale} & 63M & 77.5\% & 30 \\
 \textbf{MDEQ-small + JR (ours)} & 17M & 74.5\% & 14 \\
 \textbf{MDEQ-large + JR (ours)} & 62M & 76.8\% & 15 \\
\bottomrule
\end{tabular}}
\vspace{-.2in}
\end{table}

We additionally conduct experiments on vision tasks using the recent multiscale deep equilibrium networks (MDEQ)~\citep{bai2020multiscale}, which drive multiple feature resolutions to their equilibria simultaneously. Because of the need to maintain high- and low-resolutional feature maps at all iterative steps and generally higher channel dimensions in $f_\theta$, MDEQs are substantially slower than conventional networks like ResNets (which operate on progressively downsampled feature maps). This makes acceleration vital to broader adoption of multiscale implicit models.

The results of applying Jacobian regularization on multiscale DEQs for image classification are shown in Table~\ref{table:imagenet}. On CIFAR-10, whereas the unregularized DEQ models used 17 NFEs to reach the reported competitive level of performance, our DEQ with Jacobian regularization can converge well even within 6 iterations (in fact, we find smaller NFE values still trains, but significantly hurts generalization performance). This improvement is also obvious in Figure~\ref{subfig:cifar-convergence} and~\ref{subfig:jacobian-increase}, where we show that early stopping at threshold $T=6$ still yields good convergence with Jacobian regularization. We also demonstrate a more stable backward pass convergence throughout training in Appendix~\ref{app:extra-results}. On the much larger-scale ImageNet, where we deal with $224 \times 224$ images, the factor of reduction in NFE is not as strong (e.g., from 27 to 14 iterations, due to the receptive field issue; we'll explain this in Section~\ref{subsec:ablative-limitation}) but still yields a roughly $2\times$ acceleration. This shows that the Jacobian regularization is effective in large-scale computer vision tasks, and in the presence of multiple equilibrium points. However, we also note that as with DEQ-Transformers on WikiText-103, we notice a small slip in accuracy, which may be a result of constraining model parameterizations.

\begin{figure}[t]
\centering
\includegraphics[width=0.47\textwidth]{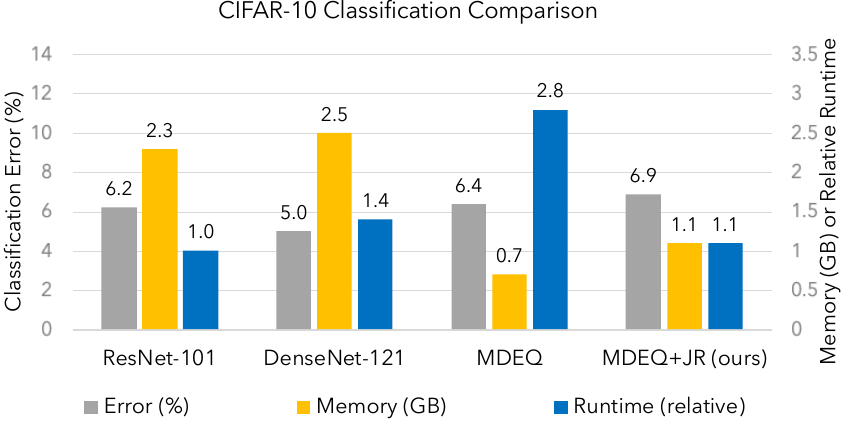}
\vspace{-.15in}
\caption{With the proposed regularization, DEQ models are competitive with popular explicit networks in accuracy, memory, and runtime. \textbf{Lower bars are better}.}
\label{fig:barchart}
\end{figure}

Figure~\ref{fig:barchart} provides a visual comparison of different models with respect to three metrics: performance, inference speed, and training memory. These are reported on the CIFAR-10 dataset. For the first time, we have an implicit-depth model that runs with a competitive level of speed and accuracy as large explicit networks such as ResNet-101, while consuming much less memory.

\subsection{Effect of Jacobian Regularization on $\rho(J_{f_\theta})$}

\begin{figure}[t]
\centering
\includegraphics[width=0.47\textwidth]{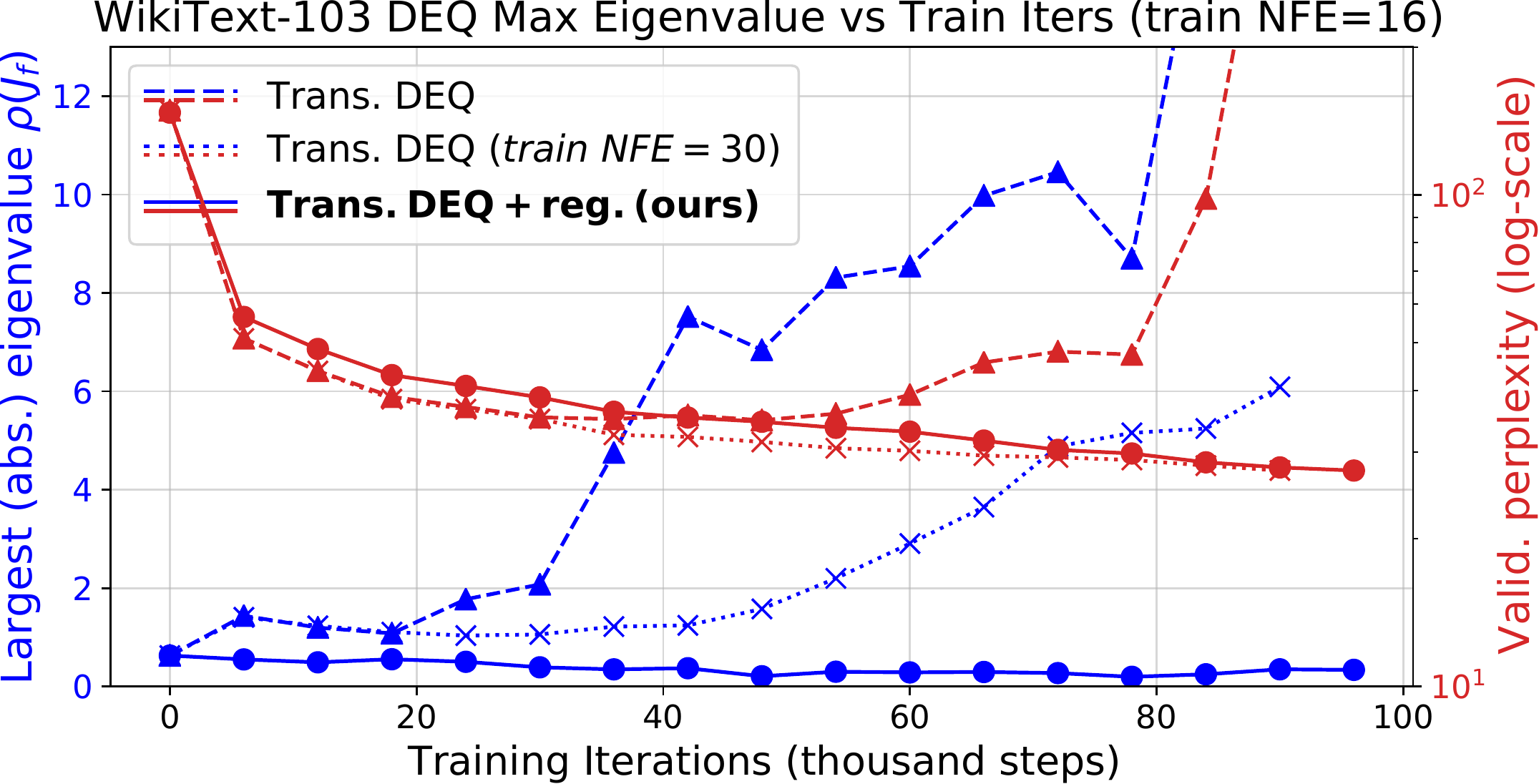}
\vspace{-.15in}
\caption{Empirical evidence of how our method constrains $\rho(J_{f_\theta})$. In contrast, insufficient NFEs (e.g., $T$=16) at training time cause a DEQ-Transformer model to explode early in the training phase.}
\label{fig:spectral-radius}
\end{figure}

In addition to the synthetic study, we also verify that the Jacobian regularization is indeed effectively constraining conditioning of $J_{f_\theta}$. Note that the underlying Jacobian matrices are large (e.g., [($B \cdot $110K) $\times$ ($B \cdot $110K)] in WikiText-103, and [($B \cdot $198K) $\times$ ($B \cdot $198K)] in ImageNet with MDEQ-small) and checking their full spectrum would be infeasible. Therefore, we conduct a study that monitors the average spectral radius $\rho(J_{f_\theta}(\mathbf{z}^\star))$ (i.e., the largest absolute eigenvalue) on the validation set, over the first 100K steps of DEQ training on WikiText-103 using the power method~\citep{mises1929praktische}; see Fig.~\ref{fig:spectral-radius}. Importantly, although $\|J_{f_\theta}\|_F$ only upper-bounds the spectral radius (see Sec.~\ref{subsec:jacobian-regularization}), we verify that our proposed regularization does effectively constrain $\rho(J_{f_\theta})$ (see \mycircle{blue}/\mycircle{red} paths in Fig.~\ref{fig:spectral-radius}), thereby making DEQs more stable. In contrast, the unregularized DEQ with the same few NFEs explodes in both eigenvalue and shortly after also in perplexity (see \mytriangle{blue}/\mytriangle{red} paths), and only works if we increase NFE to 30 (see \textcolor{blue}{$\times$}/\textcolor{red}{$\times$} paths). In general, we empirically observe that training an unregularized DEQ with insufficient NFEs generally begets extremely noisy gradients, thus leading to faster destabilization and even divergence.

\subsection{Ablative Analysis and Limitations of the Approach}
\label{subsec:ablative-limitation}

\begin{figure}[t]
\centering
\includegraphics[width=0.47\textwidth]{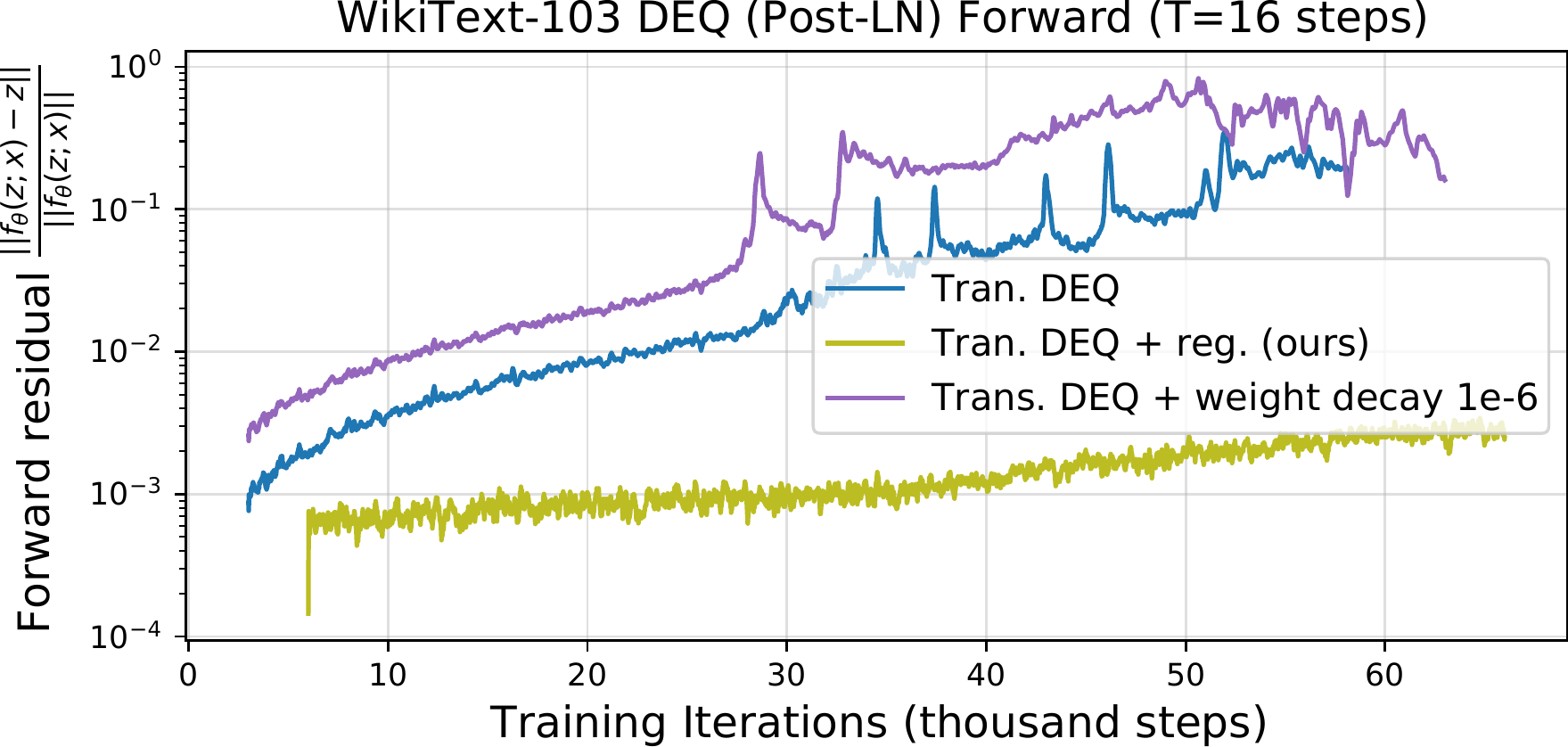}
\vspace{-.15in}
\caption{Adding weight decay of magnitude $10^{-6}$ to the DEQ-Transformer doesn't help stabilize the forward convergence.}
\label{fig:weight-decay}
\vspace{-.1in}
\end{figure}

\begin{table}[t]
\caption{Controlled experiments on the strength $\gamma$ of the Jacobian regularization. The NFE value represents the ``hard stop'' threshold we set for the corresponding DEQ models at inference.}
\label{table:cifar-ablative}
\centering
\def\arraystretch{1.02}
\vspace{2mm}
\resizebox{.48\textwidth}{!}{
\begin{tabular}{c|cccccc}
\toprule
 & NFE=1 & NFE=2 & NFE=3 & NFE=4 & NFE=5 & NFE=6 \\
 \midrule
 $\gamma=0.1$ & 82.4\% & 89.7\% & 91.9\% & 92.3\% & 92.7\% & 92.9\% \\
 $\gamma=0.6$ & \textbf{85.8}\% & \textbf{91.5}\% & \textbf{92.7}\% & \textbf{93.0}\% & \textbf{93.0}\% & \textbf{93.1}\% \\
 $\gamma=1.2$ & 84.4\% & 89.6\% & 92.2\% & 92.6\% & 92.7\% & 92.7\% \\
\bottomrule
\end{tabular}}
\vspace{-.15in}
\end{table}

We continue our discussion with some empirical ablative studies. First, while \citet{grathwohl2019ffjord} found weight decay useful for regularizing ODE-based models' NFEs, we found weight decay generally not effective in stabilizing DEQs and sometimes even counter-productive. This is illustrated in Figure~\ref{fig:weight-decay}, where after 50K steps the model started to diverge to $>500$ perplexity and stopped improving. In addition, we also conduct an ablative experiment on how the Jacobian regularization strength $\gamma$ affects the performance when we constrain NFEs to $\leq 6$ at inference time, with results shown in Table~\ref{table:cifar-ablative} (CIFAR-10 dataset). In general, we find that if $\gamma$ is too small, the final performance may be good but entails more NFEs. When $\gamma$ is too large, the accuracy does quickly converge, but the constraint imposed on the model class is too strong and eventually hurts performance (e.g., since the training loss on CIFAR-10 usually overfits to almost 0 towards the end of training, which makes the Jacobian loss dominant instead).

We also highlight two limitations of this approach. First, the addition of Jacobian regularization term does not fundamentally solve the growing instability problem, but only empirically alleviates it. This means that we have to be careful about balancing the main loss objective and this auxiliary objective (see Table~\ref{table:cifar-ablative}).
Second, while Jacobian regularization facilitates faster convergence, there are certain ``physical laws'' that we simply cannot bypass. For example, if we apply a shallow convolutional DEQ whose layer has receptive field $5 \times 5$ on a large image (e.g., $1024 \times 1024$), it is hard to be able to reach the fixed point with just 6 iterations simply because the model's receptive field may not broaden sufficiently to cover valuable context. Although one can possibly still force convergence with a large $\gamma$, it would undoubtedly hurt the performance. This explains why we need more NFEs on ImageNet than on CIFAR-10 (see Table~\ref{table:imagenet}); it also indicates that while our approach alleviates the brittleness to architectural choices, its effectiveness can still depend on the architecture. This makes global-context alternatives to ConvNets, such as self-attention-based vision layers (e.g.,ViT~\citep{dosovitskiy2020image}) likely more appealing in the implicit model setting, which we leave for future work.

\section{Conclusion}

We summarized the weaknesses of existing DEQ models, including instability \& inefficiency, architectural brittleness, and hidden memory costs. We specifically discussed the relationship between the spectral radius of the Jacobian and the stability of forward non-linear and backward linear systems of DEQ models, and provided empirical evidence of the poor conditioning of the Jacobian. This motivates our introduction of Jacobian regularization. Our experiments show that our method significantly alleviates the weaknesses of DEQs, yielding a $>2.5\times$ acceleration. This is a major step towards making implicit models more practical and suitable for large-scale real-world applications. We hope that our work will motivate further research that advances our understanding and application of this class of models.

\bibliography{regdeq}
\bibliographystyle{icml2021}

\newpage

\appendix

\twocolumn[
  \icmltitle{Accelerating Equilibrium Models by Stabilizing Their Jacobians \\ \emph{Supplementary Material}}
  \vspace{5mm}
]

\section{Dataset Information, Experimental Settings and Hyperparameters}
\label{app:experiment}

We provide below a detailed description of all tasks and settings for experiments reported in Section \ref{sec:experiments}, as well as some training specifics of the deep equilibrium network (DEQs) we use.

\subsection{1D Synthetic Dataset}

To visualize the effect of the proposed Jacobian regularization on DEQ models (see Section~\ref{sec:experiments}), we generated a synthetic dataset with 5096 pairs $(x,y)$ from the target function:
$$
y = h(x) = \frac{3}{2}x^3 + x^2 + 5x + 2\sin(x)-3+\delta
$$
where $\delta \in \mathcal{N}(0,0.05)$ are i.i.d. noise variables added to
\begin{wrapfigure}{l}{0.145\textwidth}
  \vspace{-.18in}
  \begin{center}
    \includegraphics[width=0.15\textwidth]{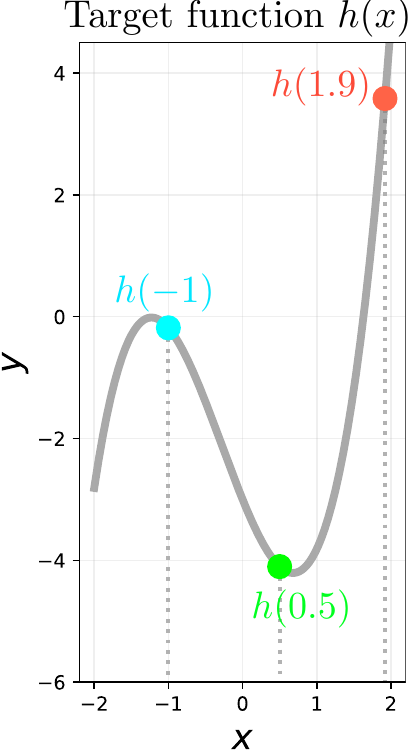}
  \end{center}
  \vspace{-.23in}
  \caption{Target function $y=h(x)$.}
  \label{appfig:target}
  \vspace{-.15in}
\end{wrapfigure}
each sample in the dataset. Specifically, we split the generated data into 4096 training samples and 1000 validation samples.

Figure~\ref{appfig:target} shows the target function. In the context of deep equilibrium networks, we aim to learn a function $z^\star(x)$ such that $z^\star = f_\theta(z^\star; x)$ and $z^\star(x) \approx h(x)$. At a high level, we should expect the \emph{intersection} between the $z_\text{out}=f_\theta(z; x)$ surface and the $z_\text{out}=z$ plane to be exactly like the gray curve in Figure~\ref{appfig:target}.

The learned DEQ equilibria $z^\star(x)$ are empirically demonstrated in Figure~\ref{fig:synthetic} in \textcolor{red}{red dashed} lines for different choices of $\gamma$. As expected, all $\gamma$ fit the target function perfectly, but the introduction of the Jacobian regularization makes the surface more flat around the fixed point.

\subsection{WikiText-103 Word-level Language Modeling}

Word-level language modeling tasks aim to predict the next word of a textual sequence by integrating the semantics and information of current and past tokens. Formally, given an input sequence $\mathbf{x}_{1:T} \in \mathbb{R}^{T \times p}$ (where $x_i \in \mathbb{R}^p$ and $T$ is the sequence length), an autoregressive sequence model $G$ produces output $G(\mathbf{x}_{1:T})=\mathbf{y}_{1:T} \in \mathbb{R}^{T \times q}$ that satisfies the causality constraint: $y_t$ depends only on $x_{1:t}$ and not on the future information $x_{t+1:T}$. When each $x_i$ represents a word (i.e., a word embedding), the task is essentially a \emph{word-level language modeling task}. This is a widely-studied problem in the NLP community (e.g.,~\citep{merity2016pointer,merityRegOpt,dai2018transformer}), and has seen practical advancement in the last few years with development of GPT-3~\citep{brown2020language,radford2019language}.

A commonly used large-scale corpus for this task is the WikiText-103~\citep{merity2016pointer} dataset, which contains 103M/217K/246K words at train/validation/test time, respectively. The entire corpus has a vocabulary size of 267K (i.e., the number of rows in the word embedding). Unlike other well-processed, much smaller datasets like Penn Treebank~\citep{Marcus93buildinga}, WikiText-103 is much more challenging as it contains many rare words and retains punctuations, numbers, upper- and lower-cases from the source Wikipedia articles; it has been the standard benchmark for many high-capacity language models in recent literature~\citep{merityRegOpt,bradbury2016quasi,dai2018transformer}. We provide a shell script in our submitted code to download this dataset.\footnote{Officially, this dataset can be downloaded at this \href{https://blog.einstein.ai/the-wikitext-long-term-dependency-language-modeling-dataset/}{link}.}

\subsection{CIFAR-10 \& ImageNet Image Classification}

The CIFAR-10~\citep{krizhevsky2009learning} dataset contains 60,000 color images of resolution $32 \times 32$ that fall into 10 object classes (with uniformly 6,000 images per class). We use the standard setting where 50K of these images are used for training and the rest 10K for validation purpose.

The ImageNet~\citep{krizhevsky2012imagenet} dataset, on the other hand, contains over 1.28M training images and 150K test images, distributed over 1,000 classes. All images are re-scaled to $224 \times 244$ resolution before they are fed into the models (as the original images are of variable resolutions and scales). This is a frequently used dataset for evaluating large-scale vision networks, and has been used for also pretraining many image feature extractor for use on downstream tasks.

For both CIFAR-10 and ImageNet, each training image goes through a canonical data augmentation process before they are fed into the model, where we perform random cropping and random horizontal flipping.

\begin{table*}[ht]
\caption{Hyperparameters, optimizer choices, and model details (at training time) for all tasks reported in Section~\ref{sec:experiments}. The arrows in the Jacobian regularization strength (e.g., $A \rightarrow B$) mean that we dynamically increase from A to B over the course of DEQ training.}
\label{table:hyperparameters}
\centering
\def\arraystretch{1.08}
\vspace{2mm}
\resizebox{.99\textwidth}{!}{
\begin{tabular}{c||c|c|c|c}
\toprule
    & \textbf{Synthetic Dataset} & \textbf{WikiText-103 language modeling} & \textbf{CIFAR-10 classification} & \textbf{ImageNet classification} \\
\midrule
\multirow{2}{*}{Architecture of $f_\theta$} & 2-Layer ReLU block & Transformer layer & Multiscale DEQ layer & Multiscale DEQ layer \\
 & (see Section~\ref{sec:experiments}) & (Pre- and Post-LN) & (residual block + fusion) & (residual block + fusion) \\
\# of Epochs & 50 & 23 & 200 & 120  \\
Batch Size & 64 & 60 & 96 & 112 \\
Optimizer & Adam & Adam & Adam & SGD \\
Start Learning rate & 0.001 & 0.00025 & 0.001 & 0.05 \\
Learning rate warmup & No & Yes, 1 epoch & No & No \\
Learning rate schedule & Cosine & Cosine & Cosine & Cosine \\
Weight Decay & 0 & 0 & 0 & $5 \cdot 10^{-5}$ \\
Hidden dimensionality & 50 & 700 (embedding size) & [28,56,112,224] (4 scales) & [32,64,128,256] \\
Input Sequence Length & N/A & 150 & N/A & N/A \\
Input Image Size & N/A & N/A & $32 \times 32$ & $224 \times 224$ \\
Normalization & None & LayerNorm~\citep{Ba2016layer} & GroupNorm~\citep{wu2018group} & GroupNorm \\
Recurrent Droput & N/A & 0.06 & 0.25 & 0.02 \\
Weight Normalization & No & Yes & Yes & Yes \\
\# of Input Injection Downsamplings & N/A & N/A & N/A & 2 \\
\midrule
Forward NFEs Threshold & 6 & 12 & 7 & 14 \\
Backward NFEs Threshold & 6 & 12 & 8 & 14 \\
Forward Threshold $\varepsilon$ & $10^{-3}$ & $10^{-3}$ & $10^{-3}$ & $10^{-3}$ \\
Backward Threshold $\varepsilon$ & $10^{-4}$ & $10^{-4}$ & $10^{-4}$ & $10^{-4}$ \\
Jacobian Reg. Strength $\gamma$ & \{0,1,2,4\} & 1.6 $\rightarrow$ 2.5 & 0.5 & 2.0 $\rightarrow$ 3.0 \\
Jacobian Reg. Frequency $p$ & 0.4 & 0.35 & 0.05 & 0.1 \\
$M$ for Hutchinson Estimator & 1 & 1 or 2 & 1 & 1 or 2 \\
\bottomrule
\end{tabular}}
\vspace{-.1in}
\end{table*}

\subsection{Training Setting and Hardware}

Our experimental protocols are intentionally set to be maximally consistent with prior work~\citep{bai2019deep,bai2020multiscale}. This includes hyperparameters (see the subsection below), other regularization methods (e.g., recurrent dropout~\citep{gal2016dropout} \& group normalization~\citep{wu2018group}), and initialization schemes (where all parameters are initialized at the start of training by sampling from $\mathcal{N}(0,0.01)$). For the multiscale DEQs that were used in the image classification task, we used 4 resolutions, where each subsequent resolution is of exactly half the height and width of the previous resolution. Although~\citet{bai2020multiscale} highlighted the need to train a ReLU-based network with softplus for stability purposes, we found it not necessary in our experiments with regularized DEQs, most likely because of the role Jacobian regularization plays in stabilizing the network convergence.

One thing to note is that empirically, rather than applying the proposed Jacobian regularization on all training iterations, we only randomly and partially apply this auxiliary loss. For example, when we set the auxiliary loss frequency $p$ to 0.5, only half of the training iterations (randomly selected) are trained with the Jacobian regularization term (see Table~\ref{table:hyperparameters}). This is motivated by the empirical observation that Jacobian-related regularizations usually hurt performance, e.g., as in its application in robust learning~\citep{hoffman2019robust}. Therefore, such partial/random supervision with the Jacobian regularization brings two benefits: 1) the rest $(1-p)$-portion of the training iterations can pick up a further speedup as we don't need to compute the Hutchinson estimator and backpropagate through it; and 2) it helps reduce the likelihood of the model overfitting on this auxiliary loss term (since, as we noted in Section~\ref{subsec:ablative-limitation}, the model could be sensitive to $\gamma$, and $M$ is small), which we generally observe to benefit the performance, though only slightly. Therefore, during training, the model would still proceed in the actual stochastic gradient direction, and only use the regularized direction occasionally.

Formally, the training objective we highlighted in Section~\ref{subsec:jacobian-regularization} should be:
\begin{align*}
\resizebox{.48\textwidth}{!}{$
\mathcal{L}_\text{total}(\mathbf{z}^\star) = \mathcal{L}_\text{orig}(\mathbf{z}^\star) + \tau \cdot \gamma \frac{\sum_{m=1}^M \|\epsilon^\top J_{f_\theta}(\mathbf{z}^\star)\|_2^2}{Md}, \ \ \epsilon_m \in \mathcal{N}(0,I_d)
$}
\end{align*}
where $\tau = \text{Bernoulli}(p)$ is a random variable and $M$ is the number of samples used for Hutchinson estimator.

All experiments in this paper, including the speed and memory benchmarks we provide, were conducted on RTX 2080 Ti GPUs. WikiText-103 language modeling and ImageNet classification models (MDEQ-small) were trained with 4 GPUs in a data-parallel setting.

\begin{table*}[t]
\caption{A more complete version of Table~\ref{table:wt103} with more memory and efficiency comparison. Memory benchmarked on batch size 15 and excludes the embedding layer. $^\dagger$ indicates unregularized model hard-stopped at inference time (while still trained with more NFEs). Overall, we find that Jacobian regularization allows us to train and predict with much fewer NFEs, at a relatively small cost in performance.}
\label{app-table:wt103}
\centering
\def\arraystretch{1.05}
\vspace{2mm}
\resizebox{.96\textwidth}{!}{
\begin{tabular}{cccccccc}
\toprule
 & Model Size & Perplexity & $t_\text{train}$ (relative) & Train NFE & Valid. NFE & Training Memory  \\
 \midrule
 AWD-Quasi RNN~\citep{bradbury2016quasi} & 159M & 33.0 & - & - & - &  7.1GB \\
 Relational Memory Core~\citep{santoro2018relational} & 195M & 31.6 & - & - & - & - \\
 Megatron-LM~\citep{shoeybi2019megatron} \textcolor{purple}{ [SOTA]} & 8300M & 10.8 & - & - & - & - \\
 Transformer-XL (18-layer)~\citep{dai2018transformer} & 110M & 24.1 & $1\times$ & - & - & 9.0GB \\
 DEQ-Transformer (Pre-LN)~\citep{bai2019deep} & 98M & \textcolor{purple}{[diverged]} & N/A & 30 & N/A & N/A \\
 DEQ-Transformer (Post-LN)~\citep{bai2019deep} & 98M & 24.0 & $3.1\times$ & 30 & 30 & 3.9GB \\
 DEQ-Transformer (Post-LN) \emph{early stopped} & 98M & 29.2 & $3.1\times$ & 30 & 12 & 3.9GB \\
 DEQ-Transformer (Post-LN)~\citep{bai2019deep} & 98M & 26.0 & $2.2\times$ & 20 & 20 & 3.6GB \\
 DEQ-Transformer (Post-LN)~\citep{bai2019deep} & 98M & \textcolor{purple}{[diverged]} & N/A & 15 & N/A & 3.6GB \\
 \midrule
 \textbf{DEQ-Transformer (Pre-LN) + JR (ours)} & 98M & \prelnppl & $1.5\times$ & 14 & 14 & 4.8GB \\
 \textbf{DEQ-Transformer (Post-LN) + JR (ours)} & 98M & \postlnppl & $1.4\times$ & 13 & 12 & 4.8GB \\
 \textbf{DEQ-Transformer (Post-LN) + JR (ours) (trained on seqlen=300)} & 98M & 23.8 & $2.2\times$ & 13 & 13 & 6.5GB \\
\bottomrule
\end{tabular}}
\end{table*}

\subsection{Hyperparameters}

We report the hyperparameters used at training time in Table~\ref{table:hyperparameters}. Except for those used in the synthetic data and for Jacobian regularization, most of the other hyperparameters were essentially taken from the original DEQ-Transformer~\citep{bai2019deep} and MDEQ~\citep{bai2020multiscale} without major modifications. For both Anderson and Broyden fixed-point solvers, we use the relative residual $\frac{\|f_\theta(\mathbf{z};\mathbf{x}) - \mathbf{z}\|}{\|f_\theta(\mathbf{z};\mathbf{x})\|}$ as a measure of convergence quality in forward and backward passes. At inference time, we generally reduce the number of NFEs (e.g., cf. Table~\ref{table:hyperparameters} and Table~\ref{table:wt103}), while the other hyperparameters (e.g., GroupNorm group sizes) are kept the same.

\begin{figure*}[t]
\centering
\begin{subfigure}[c]{0.36\textwidth}
\includegraphics[width=\textwidth]{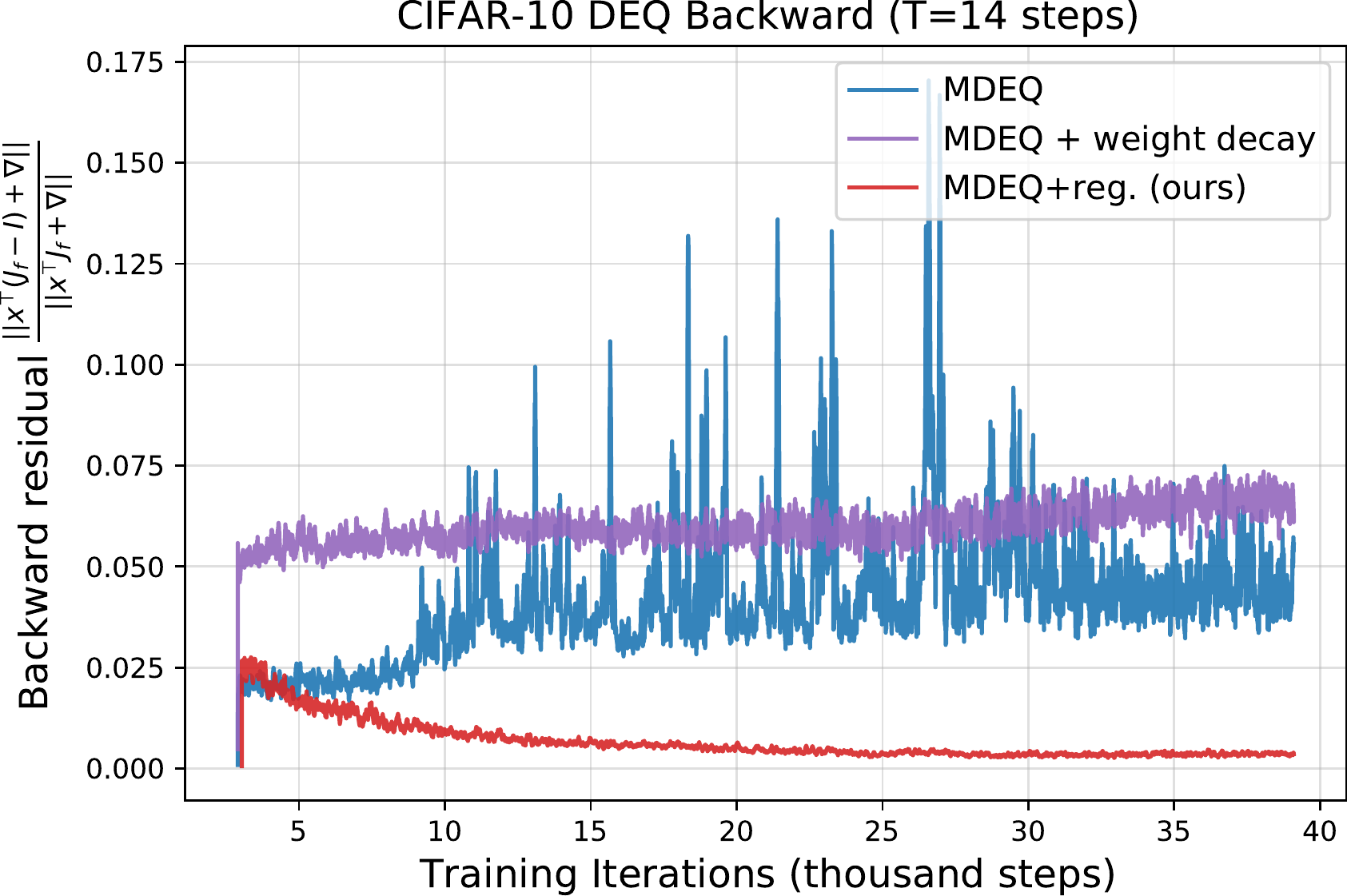}
\caption{Jacobian regularization improves both the fluctuation and quality of the backward convergence.}
\label{app-fig:cifar-backward}
\end{subfigure}
~
\begin{subfigure}[c]{0.31\textwidth}
\includegraphics[width=\textwidth]{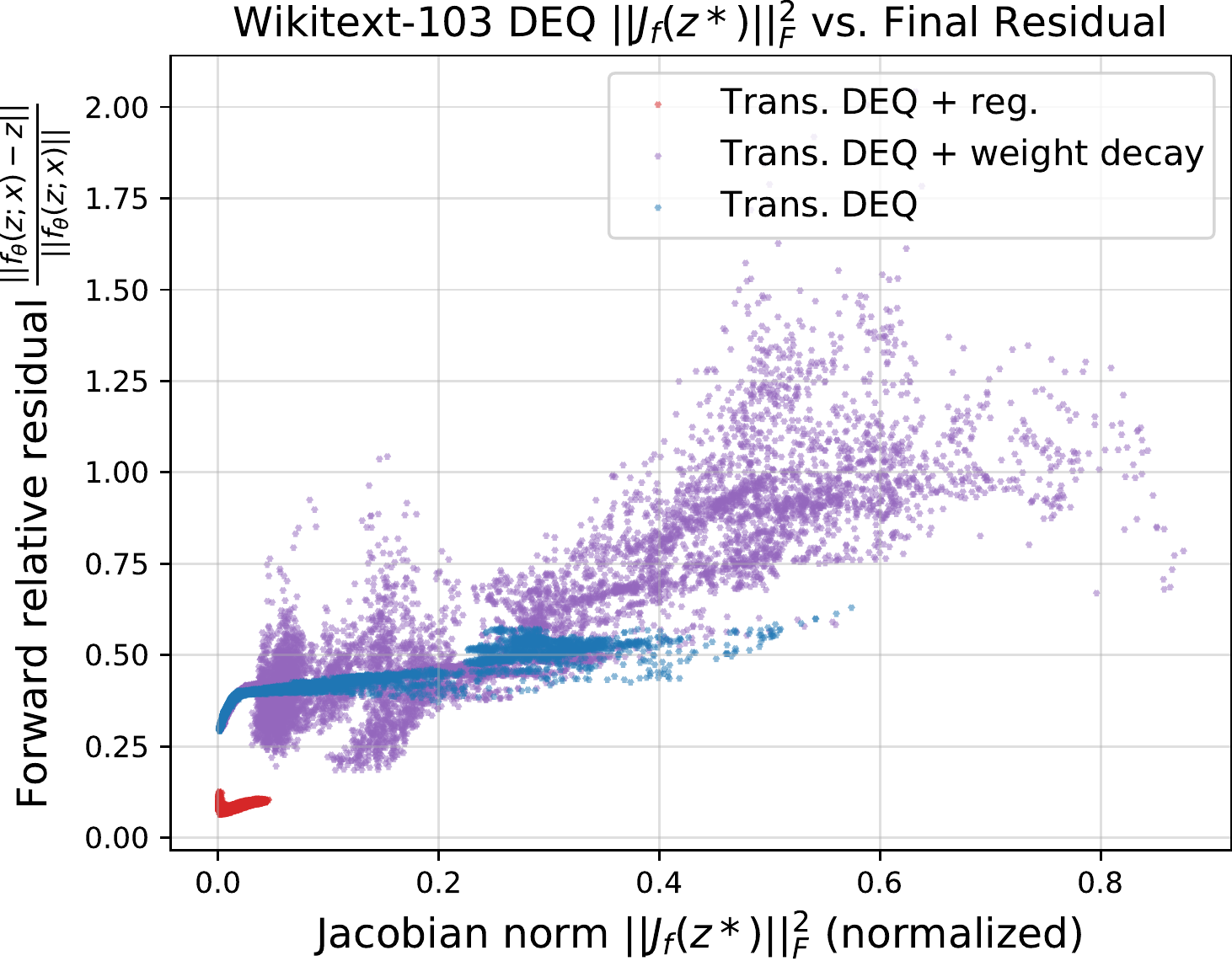}
\caption{Forward relative residual on WikiText-103 as a function of the Jacobian norm.}
\label{app-fig:wt103-jacobian-increase}
\end{subfigure}
~
\begin{subfigure}[c]{0.30\textwidth}
\includegraphics[width=\textwidth]{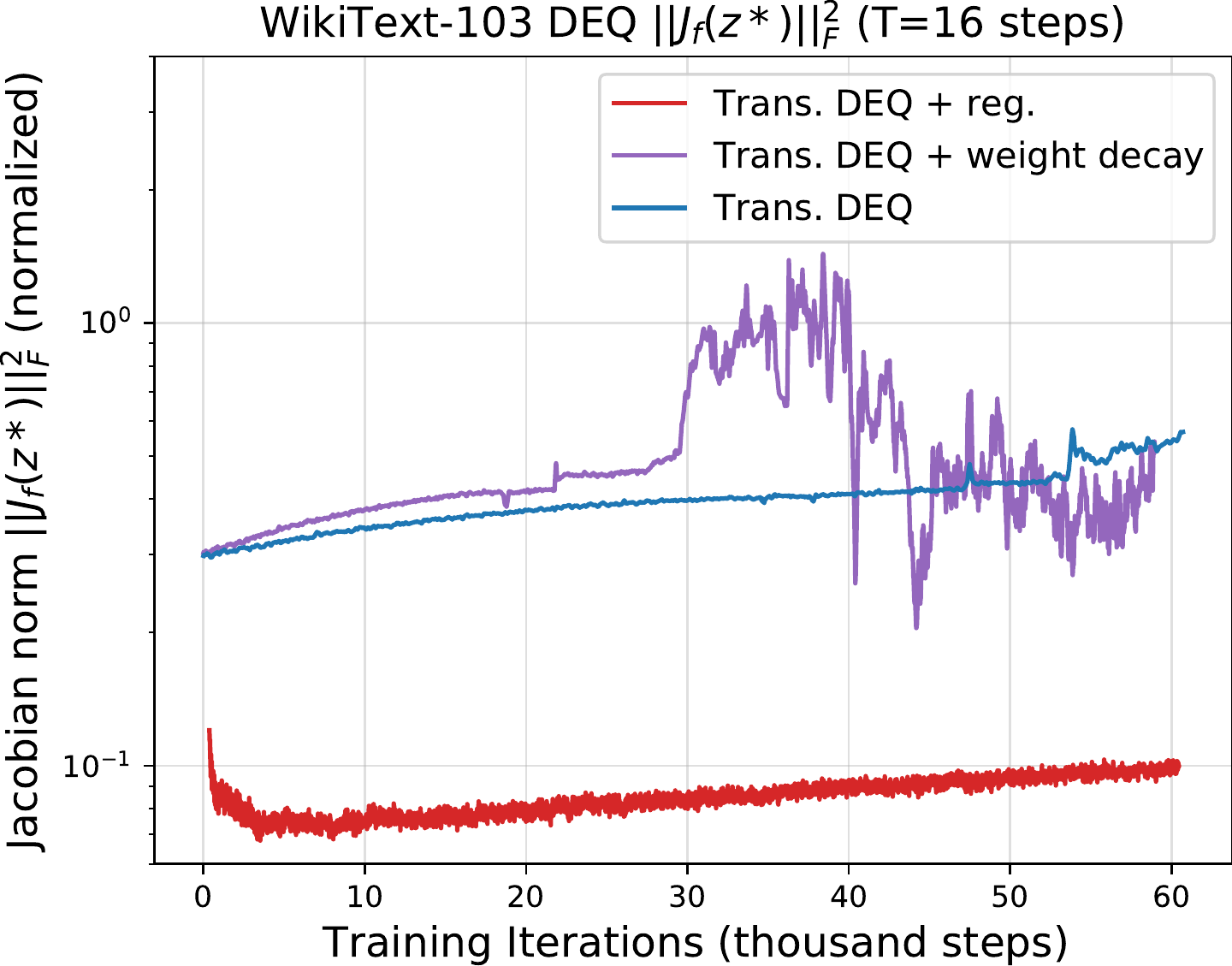}
\caption{Jacobian norm grows throughout training, even when we regularize for it.}
\label{app-fig:wt103-jacobian-time}
\end{subfigure}
\caption{Additional analysis on DEQ models' backward convergence (on CIFAR-10), Jacobian norm, etc.}
\end{figure*}

\section{Additional Experimental Results}
\label{app:extra-results}

\subsection{Memory Consumption}

As we noted in Sections~\ref{sec:regdeq} and~\ref{sec:experiments}, using Jacobian regularization and thus the vector-Jacobian-product-based Hutchinson estimator introduces some extra memory cost at training time due to the need to differentiate w.r.t. the $\|J_{f_\theta}\|_F$ term. Overall, with the same batch size and sequence length, we observe a roughly $25\%$ increase in training memory required (from about 3.9GB to 4.8GB, excluding embeddings). This is less than the memory consumption of a layer, because the reduction in NFEs needed on the other side saves the memory used by the solver (see Section~\ref{subsec:solver-choice}). However, this memory footprint is still much better than the conventional explicit Transformer-XL model, which consumes about $2\times$ as much GPU memory. With the Jacobian regularization, as we can see, the DEQ models are much more efficient in time complexity than before, while still staying competitive on the space complexity and the performance fronts.

\subsection{DEQ's Backward Convergence with Jacobian Regularization (CIFAR-10)}

As we discussed in Section~\ref{sec:regdeq}, the backward dynamics of a DEQ model is a \emph{linear} fixed point system that depends directly on the Jacobian at equilibrium (i.e., $J_{f_\theta}(\mathbf{z}^\star)$). Therefore, the backward pass stability is directly influenced by the conditioning of the Jacobian that we regularize. The stabilizing effect of the proposed Jacobian regularization on the backward pass convergence was already shown for WikiText-103 language modeling in Figure~\ref{subfig:architectural-choice-backward}, where we empirically observe that the Jacobian-regularized DEQ-Transformer's backward pass stays at a consistent level, which indicates a relatively more accurate gradient produced by the implicit function theorem.

We further corroborate this finding via empirical evidence on the CIFAR-10 dataset with a multiscale DEQ (MDEQ) instance, shown in Figure~\ref{app-fig:cifar-backward}. Compared to the original MDEQ (\textcolor{blue}{blue} line), the Jacobian-regularized version of the backward pass experiences much fewer fluctuations (and thus less stochastic gradients). We also compared to an alternative solution that uses the simple weight decay. Although it also alleviates the fluctuation problem, our empirical observations suggest that weight decay alone almost always adds more difficulty to the fixed point solving. This agrees with what we have observed in the forward pass in Section~\ref{subsec:ablative-limitation}. Such comparison can be seen in Figure~\ref{app-fig:cifar-backward} in the \textcolor{purple}{purple} line, which converged even more poorly than the original baseline after 14 backward solver iterations (with relative residual $>0.05$ and increasing slowly over training). In contrast, the regularized backward pass is more smooth and stable (\textcolor{red}{red} line) throughout training (we used $\gamma=0.5$).

\subsection{Failure of Weight Decay to Fix the Problem}

This overall inability of weight decay alone to fix the DEQ stability issue (e.g., see Figures~\ref{fig:weight-decay} and~\ref{app-fig:cifar-backward}), we believe, exactly suggests that there is a deeper \emph{implicitness} property of the model that should be regularized than just the value of individual weights. As DEQ networks typically rely on a single $f_\theta$ block, their complex non-linear structure makes their stability depend as much on the linear parts of $f_\theta$ (which weight decay does regularize) as the non-linear parts (which weight decay does not directly regularize; e.g., self-attention in $f_\theta$ if we use a Transformer layer). On the other hand, Jacobian regularization takes into account both parts as it tries to constrain the overall spectral radius of the matrix.

We also provide some additional analysis on how $\|J_{f_\theta}\|_F$ evolves during training in Figure~\ref{app-fig:wt103-jacobian-increase} and~\ref{app-fig:wt103-jacobian-time}. Specifically, even with weight decay, the convergence of DEQ-Transformer models can be quite bad (see \textcolor{purple}{purple} dots in Figure~\ref{app-fig:wt103-jacobian-increase}), with a clear correlation between the larger relative residual and larger $\|J_{f_\theta}\|_F^2$. Indeed, with a non-linear structure as complex as the multi-head self-attention, simply constraining the weights to be small is not sufficient to ensure well-conditioned Jacobians. Moreover, while the Jacobian regularization helps significantly stabilize the forward and backward convergence (see Figure~\ref{subfig:cifar-convergence},~\ref{app-fig:cifar-backward} and~\ref{fig:architectural-choice}), we note that a regularized DEQ model still in fact gradually tends to ``critical stability''. This can be seen in Figure~\ref{app-fig:wt103-jacobian-time}, where the Jacobian norm grows slowly over training iterations (\textcolor{red}{red} line) for a fixed $\gamma$, though at a rate much slower than the unregularized and weight-decayed baselines. Therefore, as we indicated in Section~\ref{subsec:ablative-limitation}, the proposed Jacobian regularization does not fundamentally \emph{fix} the growing instability problem, but only \emph{alleviates} it. This also calls for adaptive $\gamma$ scheduling during training (which we adopt in a simple form in our implementation and leave more advanced schemes for future work).

\end{document}